\documentclass{article} 

\usepackage{iclr2024_conference,times}

\usepackage{amsmath,amsfonts,bm}

\def\eqref#1{equation~\ref{#1}}

\def\1{\bm{1}}

\DeclareMathAlphabet{\mathsfit}{\encodingdefault}{\sfdefault}{m}{sl}
\SetMathAlphabet{\mathsfit}{bold}{\encodingdefault}{\sfdefault}{bx}{n}

\usepackage{hyperref}
\hypersetup{
    colorlinks=true,
    linkcolor=red,
    filecolor=magenta,      
    urlcolor=cyan,
    citecolor=mygreen2,
}
\usepackage{subcaption}
\usepackage{url}
\usepackage{graphicx}
\usepackage{booktabs}       
\usepackage{amsfonts}       
\usepackage{nicefrac}       
\usepackage{microtype}      
\usepackage{xcolor}
\definecolor{mygreen2}{RGB}{2, 142, 2}
\usepackage{amsmath}
\usepackage{multirow}
\usepackage{wrapfig}
\usepackage{enumitem}
\usepackage{float}
\usepackage{array}
\usepackage{amssymb}
\usepackage{pifont}
\usepackage{amsmath}
\newcommand{\cmark}{\ding{51}}
\newcommand{\xmark}{\ding{55}}

\usepackage{bbding}

\newcommand{\ie}{\emph{i.e., }}
\newcommand{\eg}{\emph{e.g., }}

\newcommand{\wrt}{\emph{w.r.t. }}
\newcommand{\cf}{\emph{cf. }}

\newcommand{\Mat}[1]{\textbf{#1}}

\newcommand{\Set}[1]{\mathcal{#1}}

\newcommand{\Vtr}[1]{\boldsymbol{#1}}

\definecolor{myblue}{HTML}{4B86E8}
\definecolor{myorange2}{HTML}{FE9A00}
\definecolor{mygreen}{HTML}{097969}

\newcommand{\lsh}[1]{{\color{blue}{#1}}}
\newcommand{\lshr}[1]{{\color{black}{#1}}}

\title{3D-MoLM: Towards 3D Molecule-Text Interpretation in Language Models}

\author{Sihang L$\text{i}^{1 3}$\thanks{Equal contribution.}  \And Zhiyuan Liu$^{2 *}$ \And Yanchen Lu$\text{o}^{1 3}$ \And Xiang Wan$\text{g}^{1 4}$\thanks{Correspondence to Xiang Wang and Xiangnan He. \textit{\{xiangwang1223, xiangnanhe\}@gmail.com}} \\ \AND Xiangnan H$\text{e}^{1 3 \dag}$ \And Kenji Kawaguchi$^{2}$ \And Tat-Seng Chua$^{2}$ \And Qi Tia$\text{n}^{5}$ \And \vspace{-8mm} \\ 
${}^1\!\text{ U}$niversity of Science and Technology of China \quad ${}^2\!\text{ N}$ational University of Singapore\\
${}^3\!\text{ M}$oE Key Laboratory of Brain-inspired Intelligent Perception and Cognition, USTC\\
${}^4\!\text{ I}$nstitute of Dataspace, Hefei Comprehensive National Science Center \quad ${}^5\!\text{ H}$uawei Cloud
}

\iclrfinalcopy 
\begin{document}

\maketitle

\vspace{-10pt}


\vspace{-2mm}
\begin{abstract}
    \vspace{-2mm}
    Language Models (LMs) have greatly influenced diverse domains. However, their inherent limitation in comprehending 3D molecular structures has considerably constrained their potential in the biomolecular domain. To bridge this gap, we focus on 3D molecule-text interpretation, and propose \textbf{3D-MoLM}: \underline{3D}-\underline{Mo}lecular \underline{L}anguage \underline{M}odeling. Specifically, 3D-MoLM enables an LM to interpret and analyze 3D molecules by equipping the LM with a 3D molecular encoder. This integration is achieved by a 3D molecule-text projector, bridging the 3D molecular encoder's representation space and the LM's input space. Moreover, to enhance 3D-MoLM's ability of cross-modal molecular understanding and instruction following, we meticulously curated a 3D molecule-centric instruction tuning dataset -- \textbf{3D-MoIT}. Through 3D molecule-text alignment and 3D molecule-centric instruction tuning, 3D-MoLM establishes an integration of 3D molecular encoder and LM. It significantly surpasses existing baselines on downstream tasks, including molecule-text retrieval, molecule captioning, and more challenging open-text molecular QA tasks, especially focusing on 3D-dependent properties. 
    We release our codes and datasets at \url{https://github.com/lsh0520/3D-MoLM}.
\end{abstract}
\vspace{-3mm}

\vspace{-1mm}
\section{Introduction}
\vspace{-3mm}
The advancement of Language Models (LMs) \citep{bert,gpt4,LLaMA} has triggered a series of remarkable innovations across multiple disciplines \citep{zhao2023survey}. 
Notably, LMs excel at text-based molecule understanding tasks, such as question-answering (QA) in the chemical and medical domains \citep{galactica}, by pretraining on extensive biochemical literature.
Recognizing the potential of LMs in harnessing extensive biochemical knowledge for molecule-relevant tasks,
molecule-text modeling emerges as a new research direction \citep{text2mol,molt5}. 
Previous works have been dedicated to harmonizing texts with 1D molecular sequences \citep{kvplm, galactica} and 2D molecular graphs \citep{liu2023molca, momu, stm}, aiding in tasks like molecule-text retrieval and molecule captioning.
However, they mostly leave 3D molecular structures untouched, which are crucial to understanding molecular dynamics, protein-ligand interactions, enzymatic functions, and a range of other biomolecular phenomena \citep{3D-molecule-1, 3D-molecule-2}.

\begin{figure}[ht]
\centering
\includegraphics[width=\textwidth]{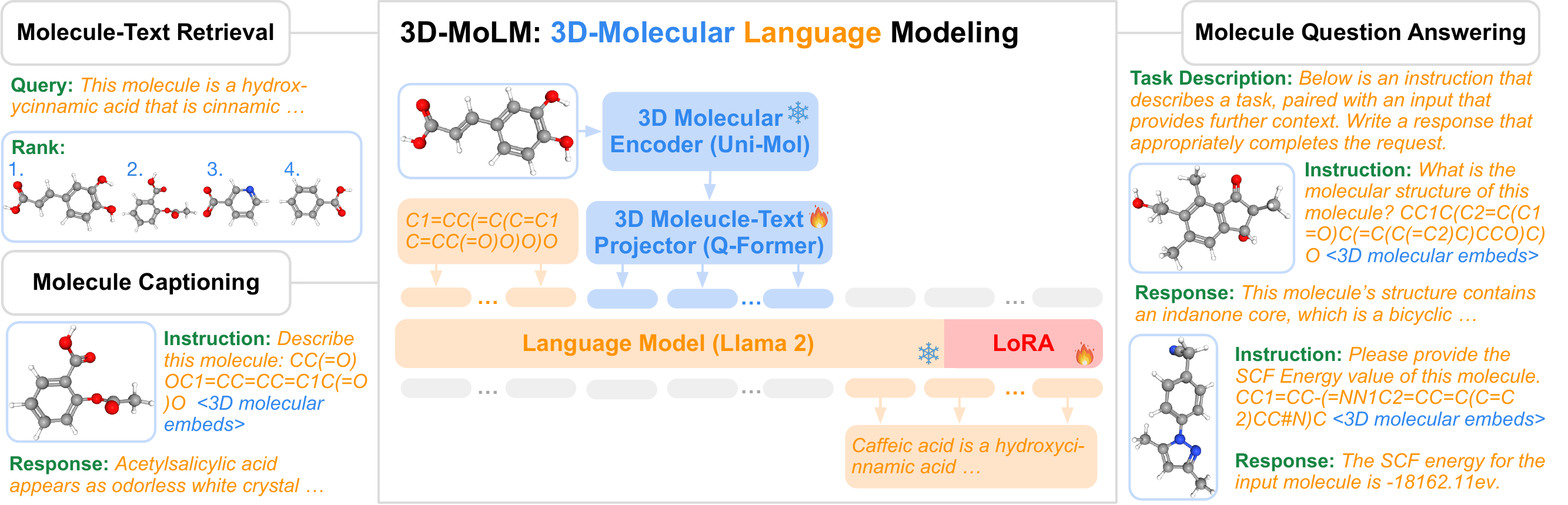}
\vspace{-16pt}
\caption{Demonstration of 3D-MoLM. 3D-MoLM is a general-purpose molecular LM that can be applied for molecule-text retrieval, molecule captioning, and molecular QA tasks. 
Flame \includegraphics[width=0.02\textwidth]{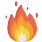} denotes tunable modules, while snowflake \includegraphics[width=0.02\textwidth]{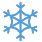} indicates frozen modules.
}
\vspace{-16pt}
\label{fig:overview}
\end{figure}

To bridge this gap, we focus on 3D molecule-text interpretation, with the goal of enabling an LM to interpret and analyze 3D molecular structures through text generation.
Given the recent successes of 3D molecular encoders in tasks like molecule property prediction, docking, and conformation prediction \citep{uni-mol,uni-mol-plus,GEM}, it is promising to incorporate one as an LM's perception module for 3D molecules. 
Upon examination of existing literature \citep{instructblip,3D-LLM,FlanModel}, 
we identify two key challenges to seamlessly integrate a 3D molecular encoder into an LM for 3D molecule-text interpretation: 
\begin{itemize}[leftmargin=*]
\item \textbf{3D Molecule-Text Alignment} maps 3D molecular representations into the input textual space where the LM can understand. 
\item \textbf{3D Molecule-centric Instruction Tuning} fine-tunes the model to follow human instructions on 3D molecule relevant tasks.
\end{itemize}




To address these challenges, we propose \textbf{3D-MoLM}: \underline{3D}-\underline{Mo}lecular \underline{L}anguage \underline{M}odeling, as depicted in Figure~\ref{fig:overview}. Specifically, it consists of two key components: 1) a 3D molecule-text projector for 3D molecule-text alignment, which aligns the latent representation spaces between the 3D molecular encoder and the LM, and 2) a dataset for \underline{3D} \underline{Mo}lecule-centric \underline{I}nstruction \underline{T}uning, \textbf{3D-MoIT}, as shown in Figure~\ref{fig:dataset}. 3D-MoIT enhances the model's ability to follow human instructions and discern 3D-dependent properties of molecules.

For 3D molecule-text alignment, we employ Q-Former \citep{blip2} as the 3D molecule-text projector, drawing inspiration from leading vision-language modeling methods \citep{minigpt4,instructblip}. 
Given a molecule's 3D structure, Q-Former converts it into tokens, which serve as 1D soft prompts~\citep{PrefixTuning}, harmonizing seamlessly with the language space of the LM. 
This translation facilitates the LM's interpretation of 3D molecular structures.   
To cultivate the Q-Former's alignment capability, two training stages are conducted  -- 
\lshr{the first stage focuses on 3D molecule-text representation learning, while the second stage optimizes for 3D molecule-text alignment. }
As depicted in Figure~\ref{fig:dataset}, these two training stages are facilitated by our collected 316K molecule-text pairs from PubChem \citep{pubchem}. 
To promote the 3D molecule-text alignment process, we manipulate the dataset by generating the 3D conformations based on SMILES using RDKit~\citep{rdkit} and enriching the molecular descriptions with GPT-3.5~\citep{ChatGPT}.
We will detail the collection and enrichment of PubChem Dataset in Section~\ref{sec:alignment} and Appendix~\ref{app:pubchem-dataset}.

Upon aligning 3D molecules with texts, we conduct instruction tuning using our curated dataset 3D-MoIT. 
It is designed to cultivate 3D-MoLM's ability to follow instructions, and to enhance its perception of 3D-dependent molecule properties. 
Specifically, 3D-MoIT is sourced from two databases: 1) PubChem, which offers a wide range of molecular properties, origins, and applications, and 2) PubChemQC \citep{pubchemqc}, which specializes in 3D-dependent molecular properties. 
As shown in Figure~\ref{fig:dataset}, for the PubChem portion, we leverage GPT-3.5 to generate QA pairs based on their descriptions.
Yet, molecular properties collected from PubChem (\eg molecular weight and LogP) can be largely inferred from 1D or 2D molecular data. 
To enhance 3D-MoIT's perception of 3D molecular structures, we further incorporate data from PubChemQC, which includes 3D-dependent molecule properties (\eg HOMO and LUMO; \cite{mcquarrie1997physical}). 
We fill these properties into a set of text templates, transforming them into instruction tuning formats, as Figure~\ref{fig:overview} illustrates.
 
Our contributions can be summarized as follows:
\begin{itemize}[leftmargin=*]
\item We propose 3D-MoLM, a new framework for 3D molecule-text interpretation. 3D-MoLM employs a 3D molecule-text projector to bridge the modality gap between a 3D molecular encoder and an LM, enabling the LM to perceive 3D molecular structures.

\item We curate 3D-MoIT, a 3D molecule-centric instruction tuning dataset. We extract and transform data from PubChem and PubChemQC to an instruction following format, to cultivate 3D-MoLM's ability in instruction following and 3D molecule-text interpretation.

\item 3D-MoLM achieves state-of-the-art performances in extensive downstream tasks. Notably, on the PubChem Dataset, for molecule-text retrieval and molecule captioning, it outperforms baselines by 20\% accuracy and 6.47 ROUGE-L, respectively. Further, it surpasses the baselines with 1D or 2D molecular perceptions on open-text QA tasks, especially on 3D-dependent properties, verifying the capability of 3D molecule-text interpretation.
\end{itemize}
\vspace{-3mm}
\section{3D-MoLM: 3D Molecular Language Modeling}
\vspace{-2mm}
\lshr{3D-MoLM incorporates a 3D molecular encoder into an LM, aiming to align 3D molecular geometries with textual concepts and facilitate a comprehensive cross-modal understanding of molecules.
Consequently, 3D-MoLM is able to read 3D molecular structures, amplifying its molecular understanding and facilitating 3D-text interpretation.
Our idea draws from related works in molecule-text modeling, multi-modal instruction tuning, and multi-modal LMs. See Appendix~\ref{sec:related-work} for a comprehensive literature review.
Here we delve into 3D-MoLM's architecture and its training pipeline. }

\begin{figure}[t]
\centering
\small
\begin{subfigure}[b]{\textwidth}
\centering
\small
\includegraphics[width=0.8\textwidth]{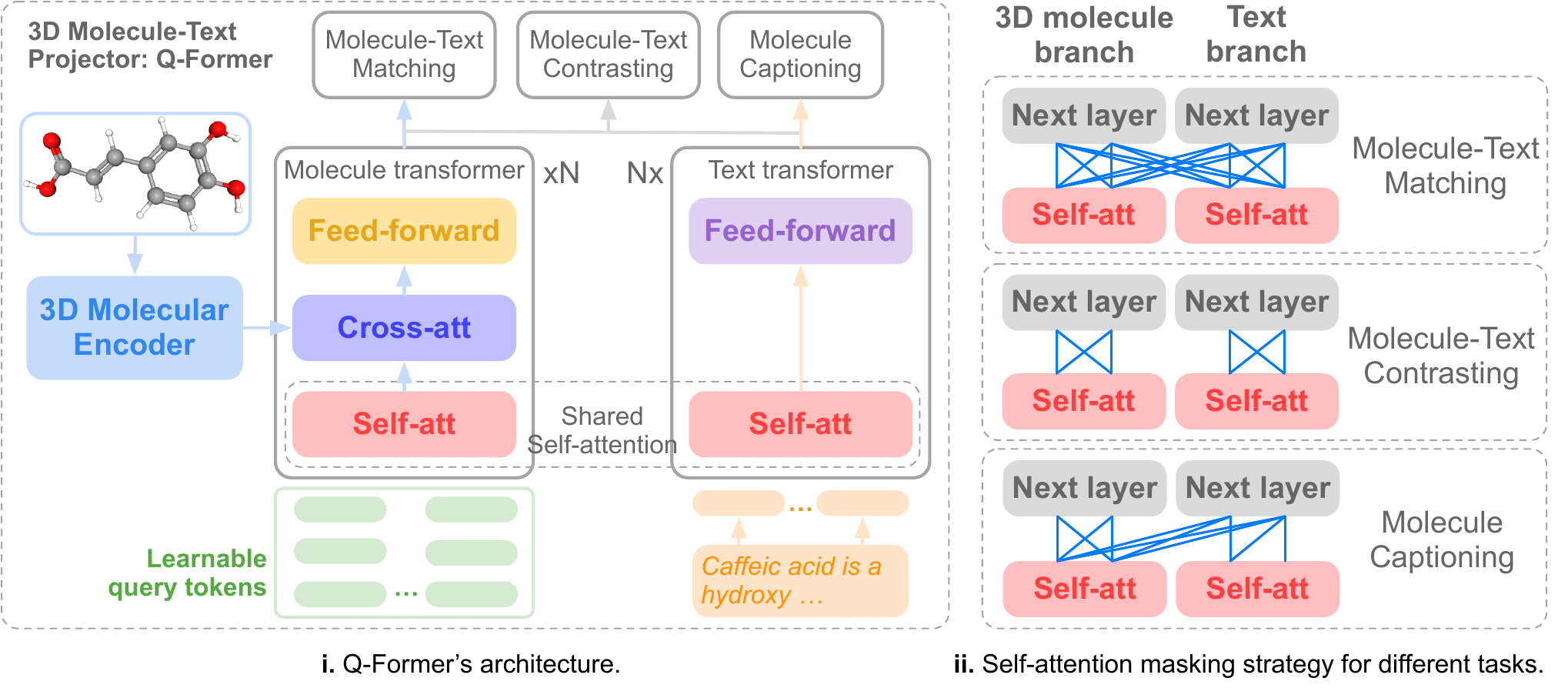}
\vspace{-4pt}
\caption{\lshr{Stage 1. The 3D molecule-text projector (\ie Q-Former) with the attached frozen 3D molecule encoder is optimized for 3D molecule-text representation learning. Stage 1 involves three training objectives: molecule-text matching, molecule-text contrasting, and molecule captioning.}}
\label{fig:qformer}
\end{subfigure}
\vspace{1cm}
\begin{subfigure}[b]{\textwidth}
\centering
\small
\includegraphics[width=0.8\textwidth]{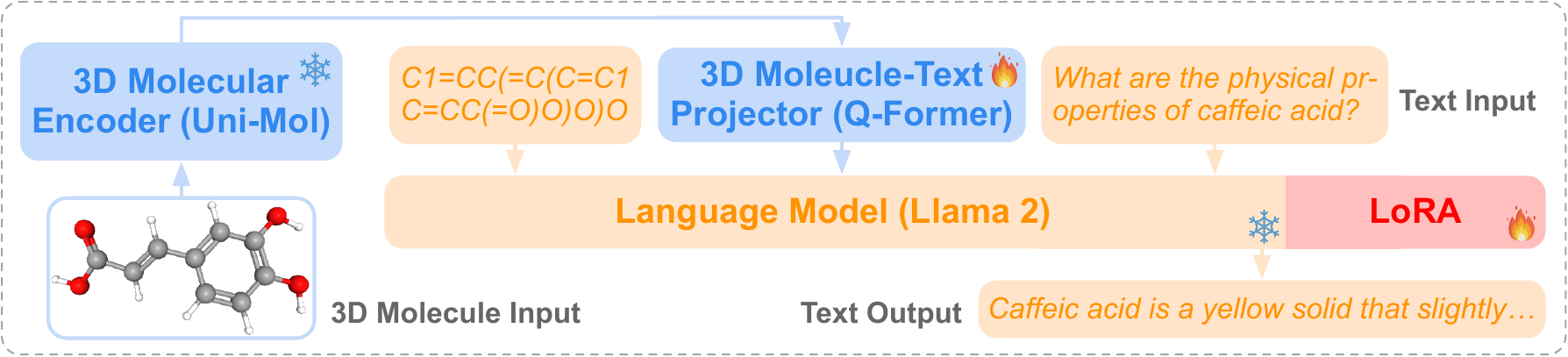}
\caption{\lshr{Stage 2 \& 3. 3D-MoLM is trained to perform 3D molecule-to-text generations given 3D molecular tokens (extracted by the Q-former) and 1D textual prompt tokens.}}
\vspace{-32pt}
\label{fig:stage23}
\end{subfigure}
\caption{\lshr{Illustration of 3D-MoLM's architectures at different stages.}}
\label{fig:qformer_and_stage23}
\vspace{-12pt}
\end{figure}

\vspace{-1mm}
\subsection{Model Architecture}
\vspace{-1mm}
3D-MoLM's architecture consists of three key components: 1) a 3D molecular encoder, focusing on encoding 3D molecular structures; 2) a 3D molecule-text projector, aiming to map the 3D molecular encoder's representations to the input space of the LM; and 3) an LM, which specializes in text generation and is later adapted for understanding 3D molecular structures.


\textbf{3D Molecular Encoder.}
We adopt Uni-Mol~\citep{uni-mol} as our 3D molecular encoder $f_{\text{mol}}$.
Specifically, Uni-Mol is pretrained on a large molecule dataset comprising 209M 3D molecular conformations.
Formally, let $m=(\Set{V},\Mat{h}, \Mat{C})$ denote a molecule, where $\Set{V}$ and $\Mat{h}$ separately represent atomic nodes and their features, and $\Mat{C} \in \space{R}^{|\Set{V}| \times 3}$ collects the 3D coordinates of nodes.
In Uni-Mol, the representation for each pair of atoms is initialized using invariant spatial positional encoding derived from 3D coordinates $\Mat{C}$. 
This encoding, grounded in the pair-wise Euclidean distances between atoms, ensures that the representation remains consistent regardless of global rotations or translations.
Subsequently, representations of atoms and atom pairs engage in a self-attention mechanism, generating the molecular representation with 3D spatial information.
Overall, the 3D molecular encoder $f_{\text{mol}}$ performs molecule encoding procedure to obtain the atomic representations:
\begin{equation}\label{molecule-encoding}
\Mat{X} = [\Vtr{x}_1, \Vtr{x}_2,...,\Vtr{x}_{|\Set{V}|}] = f_{\text{mol}}(m),
\end{equation}
where $\Vtr{x}_i$ corresponds to the representation of the $i$-th atom.

\textbf{3D Molecule-Text Projector.}
Taking inspiration from the leading vision-language models \citep{blip2, instructblip}, we architect the 3D molecule-text projector $f_{\text{pro}}$ as a Querying Transformer (\ie Q-Former) and initialize it from the Sci-BERT's checkpoint \citep{scibert}. 
As illustrated in Figure~\ref{fig:qformer}, Q-Former has two transformers \lshr{with shared self-attention layers}: one molecule transformer for processing 3D molecule features, and one text transformer for processing texts. The text transformer follows the same architecture of BERT~\citep{bert}, while the molecule transformer adds cross-attention modules between the modules of self-attention and feed-forward to extract molecule features.  
Specifically, the molecule transformer maintains $K$ learnable query tokens. Given 3D molecule input, the query tokens can interact with the 3D molecular encoder's representations through the cross-attention modules. Therefore, the $K$ query tokens' output representations contain molecule information, represented as $\Mat{M} = [\Vtr{m}_1, \Vtr{m}_2,...,\Vtr{m}_K]$.
The 3D molecule-text projector's forward function can be written as:
\vspace{-1mm}
\begin{equation}\label{cross-modal-projector}
\Mat{M} = [\Vtr{m}_1, \Vtr{m}_2,...,\Vtr{m}_K]=f_{\text{pro}}(\Mat{X}).
\end{equation}

\textbf{Language Model (LM).}
We employ Llama2 \citep{llama2} as our base LM $f_{\text{lm}}$ to leverage its powerful text generation capability and internal chemistry knowledge. Although pretrained for general-purpose usage, the extensive biomedical literature in Llama 2's pretraining corpus enables it to efficiently interpret 1D molecular sequences (\eg SMILES) and proficiently address essential QA tasks that are relevant to molecular understanding. \lshr{In this work, we let Llama2 process mixed token sequences that includes both textual tokens and 3D molecular tokens, which is detailed in Section~\ref{sec:alignment}.} Formally, we denote a mixed token sequence that include $l$ textual and molecular tokens as $\Mat{Z} = [\Vtr{z}_1, \Vtr{z}_2,...,\Vtr{z}_l]$. 
Further, the LM adopts a causal mask to generate textual response $\hat{\Mat{Z}}$ with length $n$, where the prediction for the $i$-th token, $\hat{\Vtr{z}}_i$, is dependent on its previous tokens:
\begin{equation}\label{language-model}
\hat{\Mat{Z}} = [\hat{\Vtr{z}}_{l+1}, \hat{\Vtr{z}}_{l+2},...,\hat{\Vtr{z}}_{l+n}], \qquad \hat{\Vtr{z}}_i = f_{\text{lm}}(\Mat{Z}_{<i}), \qquad \Mat{Z}_{<i} = [\Vtr{z}_1, \Vtr{z}_2,...,\Vtr{z}_{l},\hat{\Vtr{z}}_{l+1},...,\hat{\Vtr{z}}_{i-1}],
\end{equation}
where each $\hat{\Vtr{z}}_i$ is later transformed by a linear layer $f_{\text{vocab}}$ accompanied by a softmax function, converting into a probabilistic distribution over the vocabulary.
The final prediction $\tilde{\Vtr{z}}_i$ for the $i$-th token is the word in the vocabulary with the highest probability, defined as:
\begin{equation}\label{language-model-word}
\tilde{\Vtr{z}}_i=\arg \max_{w \in \text{vocab}} f_{\text{vocab}}(\hat{\Vtr{z}}_i)[w].
\end{equation}
\vspace{-2mm}

\subsection{Model Training}
\label{sec:model_training}
\begin{figure}[t]
\centering
\includegraphics[width=0.8\textwidth]{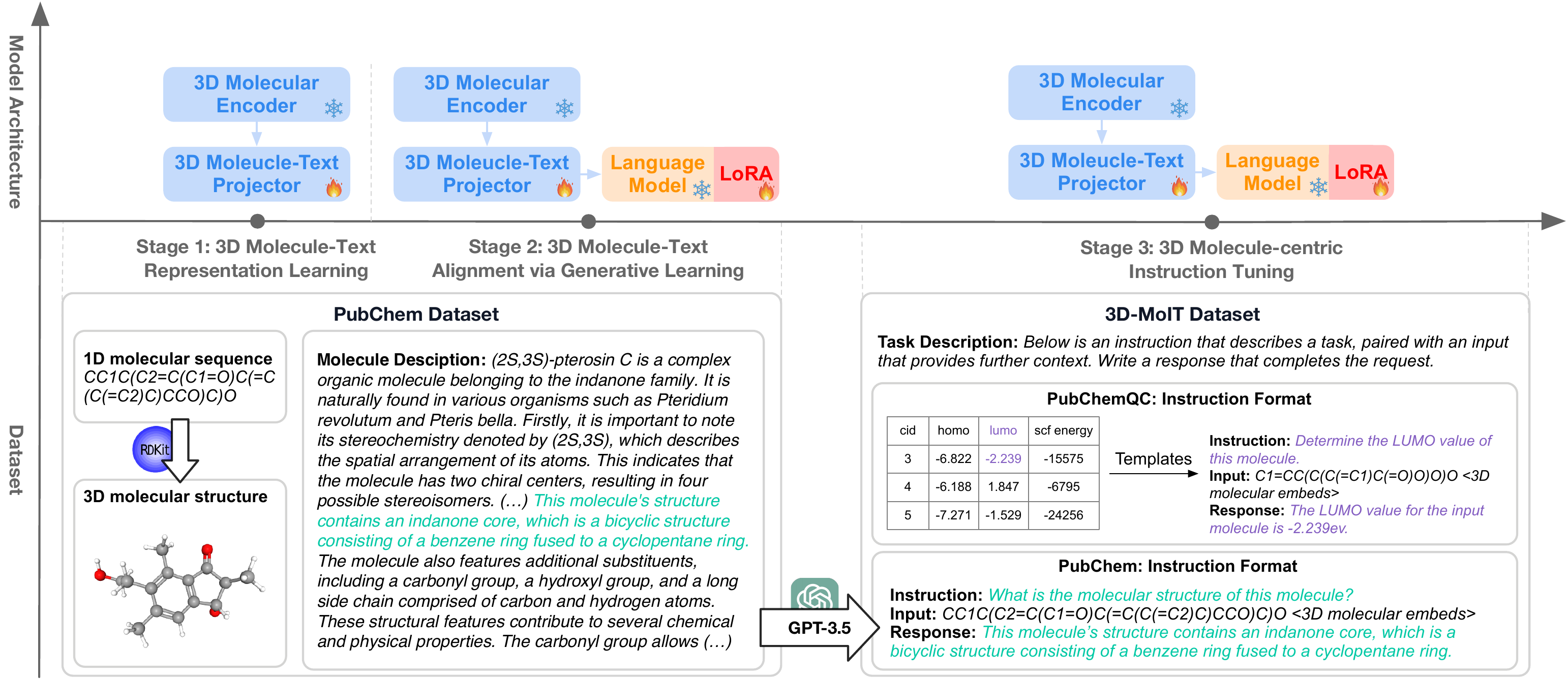}
\vspace{-4pt}
\caption{Illustration of the model architectures (upper part) and the dataset usage (bottom part) for the three training stages. \lshr{PubChem is used for the stage 1 (\ie 3D molecule-text representation learning)  and stage 2 (\ie 3D molecule-text alignment via generative learning)}. 3D-MoIT is used for 3D molecule-centric instruction tuning. Texts in the same color indicate the same information source.
}
\label{fig:dataset}
\vspace{-13pt}
\end{figure}

\vspace{-2mm}
To tackle the identified two challenges of 3D molecule-text alignment and 3D molecule-centric instruction tuning, we delineate a three-stage training pipeline (\cf Figure~\ref{fig:dataset}) for 3D-MoLM, including 1) 3D molecule-text representation learning, 2) 3D molecule-text alignment via gerative learning, and 3) 3D molecule-centric instruction tuning.

\subsubsection{3D Molecule-Text Alignment}\label{sec:alignment}
\textbf{Data Preparation -- PubChem.}
A substantial collection of 3D molecule-text pairs is essential to bridge the gap between the 3D molecular representation space and the LM's input space. 
We begin by collecting molecular SMILES-text pairs from PubChem and further employ GPT-3.5 to enrich the less annotated molecular descriptions. Molecular 3D conformations are obtained by running the MMFF algorithm in RDKit~\citep{rdkit}. 
As a result, we obtain 316K 3D molecule-text pairs for the alignment pretraining and downstream tasks of molecule-text retrieval and molecule captioning.
The details of processing, subset split, and quality evaluation are in Appendix~\ref{app:pubchem-dataset}.

\textbf{Stage 1: 3D Molecule-Text Representation Learning.} 
In the first stage, we jointly pretrain the Q-Former together with the frozen 3D molecular encoder on the collected 3D molecule-text pairs. Following BLIP-2 \citep{blip2}, we perform \lshr{multi-objective} training, including \lshr{molecule-text matching, molecule-text contrasting and molecule captioning} (\cf Figure \ref{fig:qformer}).
These training objectives are designed to cultivate Q-Former's ability to extract molecular features that resonate profoundly with the corresponding text. Specifically, the molecule-text matching task mandates the model to differentiate between matched and unmatched pairs, enhancing the fine-grained alignment of cross-modal representations.
In molecule-text contrasting, the similarity between a molecule and its corresponding text (\ie positive pair) is contrasted against those of negative pairs, aiming to maximize the mutual information between the molecular and textual representations.
\lshr{In molecule captioning}, the Q-Former is trained to generate text descriptions, based on given molecular inputs.

\lshr{\textbf{Stage 2: 3D Molecule-Text Alignment via Generative Learning.}
In this stage, we connect the 3D molecular encoder and the Q-Former with the LM for 3D molecule-to-text generation, as illustrated in Figure~\ref{fig:stage23}. The training objective is conditional language modeling: 3D-MoLM is trained to generate textual responses given 3D molecular tokens and 1D textual prompt tokens. Specifically, the 3D molecular tokens are extracted by the Q-Former, while the 1D textual prompt tokens include the molecule's SMILES sequence and a textual task description. See Appendix~\ref{app:ablation} for ablation studies on prompt templates. This training objective encourages the LM to discern the contextual interplay between textual and molecular tokens, thus aiding in 3D molecule-text interpretation. For efficiency consideration, we freeze the 3D molecular encoder and employ LoRA tuning for the LM~\citep{lora}. 
Overall, this stage aligns 3D molecular representations with the LM's textual input space, aiding in effective molecule-to-text generation.}

\begin{table}[t]
\centering
\scriptsize
\caption{Statistics of 3D-MoIT. Comp.$\rightarrow$computed property; Desc.$\rightarrow$descriptive property.}
\vspace{-8pt}
\label{tab:3d-moit-dataset}
\begin{tabular}{lccccc}
\toprule
\multirow{2}{*}{Subset}            & \multicolumn{2}{c}{PubChemQC}              & \multicolumn{3}{c}{PubChem}                                                  \\ \cmidrule(lr){2-3} \cmidrule(lr){4-6}
            & \#Mol & \#Comp. QA  & \#Mol & \#Comp. QA  & \#Desc. QA  \\\midrule
Pretrain &      3,119,717       & 12,478,868                   &      301,658       & 1,199,066                    & 1,508,290                       \\
Train    &      623,944       & 2,495,776                    &    12,000         & 46,680                       & 60,000                          \\
Valid    &       77,993      & 311,972                      &     1,000        & 3,898                        & 5,000                           \\
Test     &        77,993     & 311,972                      &       2,000      & 7,785                       & 10,000                          \\\bottomrule
\end{tabular}
\vspace{-16pt}
\end{table}

\subsubsection{3D Molecule-centric Instruction Tuning}\label{sec:instruct-tuning}
\lshr{\textbf{Stage 3: 3D Molecule-centric Instruction Tuning.}}
In the final stage, we freeze the 3D molecular encoder and conduct instruction fine-tuning to jointly optimize the 3D molecule-text projector and the LM. 
This fine-tuning has two purposes: 1) to enhance our model's ability to follow various instructions; and 2) to improve our model's understanding of 3D molecular structures, especially in recognizing 3D-dependent properties. 
It is framed as conditional text generation optimization based on the input prompt and the standard language modeling loss is adopted.
Now, we elaborate on the details of our instruction tuning dataset 3D-MoIT.

\textbf{Data Preparation -- 3D-MoIT.} 3D-MoIT sources data from the PubChem~\citep{pubchem} and PubChemQC~\citep{pubchemqc} databases. 
Specifically, the instruction tuning data from PubChem can be divided into two types: computed molecular properties and descriptive texts, where computed molecular properties are numerical values and descriptive texts characterize molecule properties in free texts. To effectively activate the 3D perception, we further include computed molecular properties from PubChemQC. PubChemQC contains 3D-dependent molecule attributes, which mostly cannot be inferred from 1D or 2D molecular representations. PubChemQC also includes DFT-determined 3D conformations, commonly regarded as the ground truth in 3D molecular modeling. The statistics of 3D-MoIT are shown in Table~\ref{tab:3d-moit-dataset}. We detail the dataset construction as follows:
\begin{itemize}[leftmargin=*]
    \item \textbf{PubChem: Instruction Format.} We select the following computed properties: molecular weight, LogP, TPSA, and complexity. They are transformed into instruction tuning format using a pre-defined set of text templates. For descriptive text, we adopt GPT-3.5 to read molecular descriptions and generate five QA pairs for each molecule, as illustrated in Figure~\ref{fig:dataset}. PubChem includes a diverse range of molecule properties, enhancing the comprehensiveness of 3D-MoIT.
    \item \textbf{PubChemQC: Instruction Format.} We select the following computed molecular properties: HOMO, LUMO, HOMO-LUMO Gap, and SCF-energy. These properties are transformed into instruction tuning format by filling the values into pre-defined text templates, as illustrated in Figure~\ref{fig:dataset}. We use the processed dataset released by~\citep{molecule3d} and follow the scaffold split.
\end{itemize}

\vspace{-0.2cm}
\section{Experiment} \label{sec:experiment}
\vspace{-0.2cm}
In this section, we conduct extensive experiments, including molecule-text retrieval, molecule captioning, and open-text molecular QA tasks, to demonstrate the effectiveness of 3D-MoLM for 3D molecule-text interpretation.
See Appendix~\ref{app:exp-setup} for experimental details of each downstream task.

\vspace{-0.1cm}
\subsection{Molecule-Text Retrieval}\label{sec:retrieval}
\vspace{-0.1cm}
We assess the Stage-1 checkpoint of 3D-MoLM on the downstream subsets of PubChem Dataset for molecule-text retrieval. 
These subsets contain real-world molecules paired with textual descriptions longer than 20 words. 
We opt not to evaluate existing molecule-text datasets of PCDes~\citep{kvplm}, because of data leakage. PCDes and our curated PubChem dataset stem from the same source (\ie some PCDes test data present in our pretraining set).
We employ Sci-BERT~\citep{scibert}, KV-PLM~\citep{kvplm}, and MoMu~\citep{momu} as baselines and evaluate the performance by Accuracy and Recall@20, both within a batch of 64 samples and across the entire test set. 
Baselines are initiated from their official checkpoints and finetuned using the downstream partition of the PubChem Dataset, except $\text{MoMu}^\dag$, which is our re-implementation with the original PubChem texts without GPT-3.5 enrichment.
From Table~\ref{tab:cross-modal-retrieval}, we have the following observations:

\begin{table}[t]
\centering
\scriptsize
\caption{Molecule-Text retrieval results on the PubChem Dataset. 
$\dag$ denotes pretraining on the original PubChem texts without GPT-3.5 enrichment. 
We report performances of both using molecule to retrieve text (M2T) and using text to retrieve molecule (T2M). 
}
\vspace{-3mm}
\label{tab:cross-modal-retrieval}
 \begin{tabular}{lcccccccc}
    \toprule
    & \multicolumn{4}{c}{Retrieval in batch} & \multicolumn{4}{c}{Retrieval in test set} \\ 
    & \multicolumn{2}{c}{M2T (\%)} & \multicolumn{2}{c}{T2M (\%)} & \multicolumn{2}{c}{M2T (\%)} & \multicolumn{2}{c}{T2M (\%)} \\\cmidrule(lr){2-3}\cmidrule(lr){4-5} \cmidrule(lr){6-7} \cmidrule(lr){8-9}
    Model & Acc & R@20 & Acc & R@20 & Acc & R@20 & Acc & R@20 \\\midrule
    \multicolumn{2}{l}{\textbf{1D SMILES}} & \multicolumn{1}{l}{} & \multicolumn{1}{l}{} & \multicolumn{1}{l}{} & \multicolumn{1}{l}{} & \multicolumn{1}{l}{} & \multicolumn{1}{l}{} & \multicolumn{1}{l}{} \\
    Sci-BERT & 85.32 & 98.74 & 84.20 & 98.43 & 41.67 & 87.31 & 40.18 & 86.77 \\
    KV-PLM & 86.05 & 98.63 & 85.21 & 98.47 & 42.80 & 88.46 & 41.67 & 87.80 \\\midrule
    \multicolumn{2}{l}{\textbf{2D Graph}} & \multicolumn{1}{l}{} & \multicolumn{1}{l}{} & \multicolumn{1}{l}{} & \multicolumn{1}{l}{} & \multicolumn{1}{l}{} & \multicolumn{1}{l}{} & \multicolumn{1}{l}{} \\
    MoMu-S & 87.58 & 99.24 & 86.44 & 99.38 & 47.29 & 90.77 & 48.13 & 89.92 \\
    MoMu-K & 88.23 & 99.41 & 87.29 & 99.42 & 48.47 & 91.64 & 49.46 & 90.73 \\
    \lshr{$\text{MoMu-S}^\dag$} & \lshr{90.43} & \lshr{99.53} & \lshr{89.38} & \lshr{99.60} & \lshr{60.51} & \lshr{93.24} & \lshr{58.36} & \lshr{91.35} \\
    \lshr{$\text{MoMu-K}^\dag$} & \lshr{90.89} & \lshr{99.67} & \lshr{90.16} & \lshr{\underline{99.44}} & \lshr{62.07} & \lshr{93.06} & \lshr{59.17} & \lshr{92.01} \\
    \midrule
    \multicolumn{2}{l}{\textbf{3D Conformation}} & \multicolumn{1}{l}{} & \multicolumn{1}{l}{} & \multicolumn{1}{l}{} & \multicolumn{1}{l}{} & \multicolumn{1}{l}{} & \multicolumn{1}{l}{} & \multicolumn{1}{l}{} \\
    $\text{3D-MoLM}^\dag$ & \textbf{94.48} & \underline{99.74} & \textbf{94.78} & 99.34 & \textbf{72.06} & \textbf{96.42} & \textbf{71.30} & \textbf{95.96} \\
    3D-MoLM & \underline{93.50} & \textbf{100.00} & \underline{92.89} & \textbf{99.59} & \underline{69.05} & \underline{95.91} & \underline{70.13} & \underline{94.88}\\
    \bottomrule
    \end{tabular}
    \vspace{-12pt}
\end{table}

1) 3D-MoLM surpasses existing baselines, including both molecular 1D-language models (\ie Sci-BERT, KV-PLM) and 2D-language models (\ie MoMu-S/K), by a large margin.
The improvement can be attributed to two pivotal factors. 
\lshr{
Firstly, the performance gain $\text{3D-MoLM}^\dag$ over $\text{MoMu}^\dag$, which are both pretrained on the same PubChem molecule-text pairs, demonstrates that Q-Former benefits from multi-objective pretraining across diverse granularities, distinguishing it from other models that predominantly rely on the molecule-text contrastive learning objective. 
Secondly, the scale of the PubChem Dataset, which we curated, offers a considerable advantage, which is verified by the performance gain $\text{MoMu}^\dag$ over the original MoMu.
With the collection of 301K molecule-text pairs for pretraining, it markedly surpasses the 15K pairs in the MoMu \wrt scale.
This increased scale provides our model with a richer and more diverse set of training examples, allowing it to better capture the molecular-text interactions.
}
Such superior performance underscores the effectiveness of 3D molecular representation learning in Stage 1, demonstrating the capability of Q-Former to extract molecular features that are strongly related to the textual descriptions.

2) The retrieval performance on the PubChem test set appears to be negatively impacted by GPT-3.5 enrichment. 
We infer that this decline is caused by the enrichment process enlarging the distribution gap between the pretraining and downstream subsets. 
While the original texts might be succinct and hence less informative, they exhibit word patterns more congruent with those in the test set, in stark contrast to the distinctive GPT-style exhibited by the enriched texts.
Nonetheless, we argue that this enrichment process benefits forging a more comprehensive connection between molecular structures and their inherent properties.
This assertion is further substantiated by subsequent experiments of textual generation tasks, where GPT-3.5 enrichment boosts the performance of 3D-MoLM.

\newcolumntype{M}[1]{>{\scriptsize\arraybackslash}m{#1}}
\begin{table*}[t]
\centering
\caption{\lshr{Molecule captioning results on PubChem Dataset.
$\dag$ denotes pretraining on the original PubChem texts without GPT-3.5 enrichment.
Llama2-7B, without a molecule-text projector, goes through Stage 2 training with the prompt of 1D SMILES.
2D-MoLM replaces the 3D molecular encoder with a 2D molecular encoder and goes through the same training process as 3D-MoLM.
}
}
\vspace{-6pt}
\scriptsize
\begin{subtable}[t]{\textwidth}
\centering
\caption{Molecule captioning results.}
\vspace{-4pt}
\begin{tabular}{llcccccc}\toprule
    Type & Model                  & BLEU-2               & BLEU-4               & ROUGE-1              & ROUGE-2              & ROUGE-L              & METEOR               \\\midrule
    \multirow{11}{*}{Specialist} &
    \multicolumn{2}{l}{\textbf{1D SMILES}}            & \multicolumn{1}{l}{} & \multicolumn{1}{l}{} & \multicolumn{1}{l}{} & \multicolumn{1}{l}{} & \multicolumn{1}{l}{} \\
    &MolT5-Small                                & 22.53                 & 15.23                 & 30.44                 & 13.45                 & 20.30                 & 23.98                 \\
    &MolT5-Base                                 & 24.51                 & 16.61                 & 32.19                 & 14.04                 & 21.35                 & 26.10                 \\
    &MolT5-Large                                 & 25.87                 & 17.28                 & 34.07                 & 16.42                 & 23.41                 & 28.04                 \\
    \cmidrule{2-8}
    &\multicolumn{3}{l}{\textbf{1D SMILES + 2D Graph}} & \multicolumn{1}{l}{} & \multicolumn{1}{l}{} & \multicolumn{1}{l}{} & \multicolumn{1}{l}{} \\
    &MoMu-Small                                  & 22.86                 & 16.01                 & 30.98                    & 13.65                    & 20.75                 & 24.35                 \\
    &MoMu-Base                                  & 24.74                 & 16.77                 & 32.45                    &14.62                    & 22.09                 & 27.16                 \\
    &MoMu-Large                                 & 26.34                 & 18.01                 & 34.75                    & 16.86                    & 24.76                 & 28.73                 \\
    \cmidrule{2-8}
    &\multicolumn{3}{l}{\textbf{1D SMILES + 3D Conformation}}  & \multicolumn{1}{l}{} & \multicolumn{1}{l}{} & \multicolumn{1}{l}{} & \multicolumn{1}{l}{} \\
    &3D-MoLM$\dag$    & \underline{29.82}        & \underline{22.39}        & \textbf{37.23}        & \underline{22.49}        & \underline{31.07}        & \underline{32.69}       \\
    &3D-MoLM    & \textbf{30.32}        & \textbf{22.52}       & \underline{36.84}        & \textbf{22.32}       & \textbf{31.23}        & \textbf{33.06}       \\      
    \midrule
    \multirow{7}{*}{\lshr{Generalist}} 
    &\multicolumn{3}{l}{\textbf{\lshr{1D SMILES}}}  & \multicolumn{1}{l}{} & \multicolumn{1}{l}{} & \multicolumn{1}{l}{} & \multicolumn{1}{l}{} \\
    &\lshr{Llama2-7B}    & \lshr{27.01}        & \lshr{20.94}        & \lshr{35.76}        & \lshr{20.68}        & \lshr{28.88}        & \lshr{32.11}       \\
    \cmidrule{2-8}
    &\multicolumn{3}{l}{\textbf{\lshr{1D SMILES + 2D Graph}}}  & \multicolumn{1}{l}{} & \multicolumn{1}{l}{} & \multicolumn{1}{l}{} & \multicolumn{1}{l}{} \\
    &\lshr{2D-MoLM}    & \lshr{27.15}        & \lshr{21.19}       & \lshr{36.02}        & \lshr{20.76}       & \lshr{29.12}        & \lshr{32.28}       \\
    \cmidrule{2-8}
    &\multicolumn{3}{l}{\textbf{\lshr{1D SMILES + 3D Conformation}}}  & \multicolumn{1}{l}{} & \multicolumn{1}{l}{} & \multicolumn{1}{l}{} & \multicolumn{1}{l}{} \\
    &\lshr{3D-MoLM$\dag$}    & \lshr{\textbf{29.25}}        & \lshr{\textbf{22.07}}       & \lshr{\underline{36.48}}        & \lshr{\textbf{21.80}}       & \lshr{\textbf{30.95}}        & \lshr{\underline{33.12}}       \\
    &\lshr{3D-MoLM}    &     \lshr{\underline{28.95}}        & \lshr{\underline{21.63}}        & \lshr{\textbf{36.51}}        & \lshr{\underline{21.26}}        & \lshr{\underline{30.02}}        & \lshr{\textbf{33.55}}       \\ 
    \bottomrule
    \addlinespace[0.1cm]
    \end{tabular}
\end{subtable}

\begin{subtable}[t]{\textwidth}
\centering
\caption{A sample of molecule captioning.}
    \begin{tabular}{M{2.25cm} M{5cm} M{5cm}}
        \toprule
        \scriptsize{Molecule} & Ground Truth & 3D-MoLM  \\
        \midrule
        \includegraphics[width=0.17\textwidth]{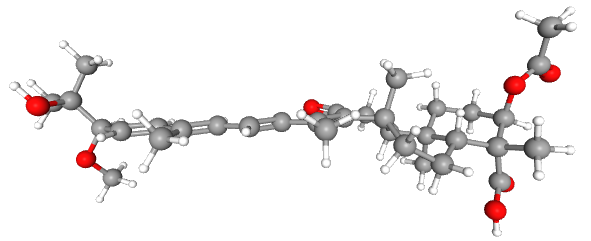} & 
        \lsh{Globostellatic acid} C is a \lsh{tricyclic triterpenoid} of the isomalabaricane group. \lsh{It has a role as} an antineoplastic agent and \lsh{a metabolite}. \lsh{It is a tricyclic triterpenoid}, \lsh{an acetate ester}, an enone, an ether and an oxo \lsh{monocarboxylic acid}. It is a conjugate acid of a globostellatate C(1-). & 
        \lsh{Globostellatic acid} B is a \lsh{tricyclic triterpenoid} with formula C34H48O6, originally isolated from Globostellata. \lsh{It has a role as a metabolite}. \lsh{It is a tricyclic triterpenoid}, \lsh{a monocarboxylic acid} and \lsh{an acetate ester}. Globostellatic acid B is a natural product found in Globostellata with data available. \\
        \bottomrule
    \end{tabular}
\end{subtable}
\label{tab:molecule-captioning}
\vspace{-20pt}
\end{table*}

\vspace{-2mm}
\subsection{Molecule Captioning}\label{sec:captioning}
\vspace{-1mm}
\lshr{\textbf{Generalist and Specialist Models.} For generation tasks of molecule captioning and open-text molecule QA, we present 3D-MoLM's performances of two different variants: the specialist, which is fine-tuned for each generation task separately, and the generalist, which is a unified model trained on all the generation tasks. Both models are obtained by fine-tuning the Stage-2 checkpoint.
}

Here we evaluate 3D-MoLM for molecule captioning on the PubChem Dataset. 
We opt not to use CheBI-20~\citep{molt5} for two reasons: firstly, its origin from PubChem poses potential data leakage issues, akin to the concerns with PCDes; secondly, CheBI-20 is curated in a way that molecular names are replaced with ``the molecule'', driving the model to emphasize properties instead of names. 
However, the molecular nomenclature system is intrinsically related to the identification of distinct molecular structures, encompassing features such as hydrocarbon chains and benzene rings. 
Consequently, a model's ability to accurately predict these names serves as a testament to its adeptness in comprehending the underlying molecular structures.
Thus, we elevate the task's complexity by retaining molecular names in the texts, positioning this task as a combination of molecule captioning without molecular names~\citep{molt5}, and name prediction~\citep{iupac-name}.
\lshr{To demonstrate the effectiveness of 3D molecular perception, we include 3D-MoLM's variants of 1D (\ie Llama2-7B) and 2D (\ie 2D-MoLM) perception as baselines.
Specifically, Llama2-7B, without a molecule-text projector, goes through Stage 2 \& 3 training using 1D SMILES as molecule representations.
2D-MoLM replaces the 3D molecular encoder with a 2D molecular encoder~\citep{graphmvp}, and undergoes the same training process as 3D-MoLM.
Specialist models are fine-tuned using the training set from the PubChem Dataset. }
Table~\ref{tab:molecule-captioning} presents the performances with metrics of BLEU, ROUGE, and METEOR, accompanied by a concrete sample.
We observe that:

1) 3D-MoLM demonstrates superior performance across the board, with the highest scores on all evaluation metrics. 
While slightly erring in identifying the molecule as ``Globostellatic acid B'' rather than ``Globostellatic acid C'', it pinpoints roles and primary structures such as the tricyclic triterpenoid, acetate ester, and monocarboxylic acid.
This demonstrates the effectiveness of 3D molecule-text alignment training to bridge the gap between 3D molecular representations and LM's input space.
\lshr{We also provide detailed analysis and discussion on failure cases in Appendix \ref{app:failure-case}.}
 
2) The enrichment process via GPT bolsters the text generation capability based on 3D molecular structures. 
This underscores the hypothesis that the enrichment strengthens the connection between molecular structures and their associated properties, enhancing cross-modal molecular understanding.
Similar gain can be observed in the following open-text QA tasks as well.

3) 3D-MoLM's heightened performance, when juxtaposed with finetuned Llama2-7B and 2D-MoLM subjected to a similar training process but modeling molecules as 1D SMILES and 2D graphs, highlights the pivotal role of 3D structure perception in bolstering molecular understanding.

\subsection{Open-text Molecular Question-Answering (QA)}\label{sec:open-text-qa}


We evaluate 3D-MoLM for open-text molecular QA on the 3D-MoIT dataset. 
Considering that open-text molecular QA is mostly unexplored in existing works, we mainly compare 3D-MoLM with its variants of 1D or 2D molecular perceptions. \lshr{Notably, we report performances of specialists, which are trained for each task separately, and generalists, which are unified models trained on all the generation tasks.} Table~\ref{tab:open-text-qa} presents the quantitative evaluations and QA samples. We observe that:

\textbf{Observations for Descriptive Property QA:} 1) Superiority of 3D-MoLM over baselines. 
It exhibits a commendable performance in 3D molecule-text interpretation, clearly surpassing the baselines. 
Specifically, It correctly identifies beryllium acetate as a solid that is soluble in water and goes beyond the ground truth by providing supplementary details, such as pinpointing the elevated melting and boiling points and attributing them to the potent intermolecular forces between beryllium and acetate, which underlines its depth of molecular understanding.

2) Benefit of descriptive instructions. 
Incorporating them amplifies the model's molecular comprehension. 
This can be verified by that through instruction tuning, Llama2-7B (generalist) advances over its initialization point, manifesting improvements of 2.46 in BLEU-2 and 3.22 in METEOR.

\textbf{Observations for Computed Property QA:} 1) 3D-MoLM achieves superior performances on computed property QA task, consistently achieving the lowest MAE, especially on those properties intrinsically determined by 3D conformations (\ie highlighted properties in Table \ref{tab:computed-qa}).
A remarkable performance lead of 0.77 eV among generalists on HOMO-LUMO Gap accentuates 3D-MoLM's adeptness at 3D molecular understanding, which we attribute to its 3D perception. 
However, for properties that mainly stem from atomic compositions and interatomic connections (\ie molecule weight, LogP, TPSA, and complexity), the advantage, while persistent, is more subdued. 
This aligns with Uni-Mol, which displays larger advantages over 2D molecular models for predicting quantum mechanical properties grounded in 3D structures.

2) Efficacy of instruction tuning. 
Instruction tuning amplifies the model's capability to adhere to and act upon instructions. 
This is illuminated by Llama2-7B's official checkpoint's occasional ineptitude in generating valid numerical responses, particularly when tasked with approximating 3D-dependent computed properties that aren't straightforwardly derivable from SMILES.

\lshr{
3) Comparison with Uni-Mol. 3D-MoLM can enhance the accuracy of molecular property prediction by leveraging both the rich contextual knowledge found in chemistry literature and 3D molecular conformations. 
For instance, the pretraining corpus of 3D-MoLM contains descriptions of hydrophobicity (LogP) and solubility (TPSA). 
While Uni-Mol excels at predicting molecular properties by interpreting 3D conformations, it cannot utilize textual descriptions of chemical properties. 
This dual-source knowledge utilization can potentially enhance the prediction of molecular properties.
}

\lshr{\textbf{Observations for generalist and specialist:} While the generalist model slightly underperforms in comparison to the specialist models, it still exhibits a performance gain over other baselines. This underscores 3D-MoLM's versatility and capability to effectively handle multiple tasks.}

\begin{table*}[t]
\centering
\caption{Open-text QA results on 3D-MoIT.
* denotes the official checkpoint without any finetuning.
$\dag$ denotes molecule-text alignment on the original PubChem texts without GPT-3.5 enrichment.
Llama2-7B, without a molecule-text projector, goes through Stage 3 instruction tuning by modeling molecules as 1D SMILES.
2D-MoLM replaces the 3D molecular encoder with a 2D molecular encoder and goes through three-stage training as 3D-MoLM.
}
\scriptsize
\vspace{-4pt}
\begin{subtable}[t]{\textwidth}
\centering
\caption{Descriptive property QA results.}
\vspace{-4pt}
\begin{tabular}{llcccccc}
\toprule
Type& Model &  BLEU-2 & BLEU-4 & ROUGE-1 & ROUGE-2 & ROUGE-L & METEOR \\
\midrule
\multirow{4}{*}{Specialist} 
&Llama2-7B        &  28.15 & 23.24 & 35.14 & 22.08  &  30.41 & 46.87 \\
&2D-MoLM        & \underline{30.84} & \underline{25.09} &  38.46 & \underline{24.22}  & \underline{33.04}  & 50.92 \\
&3D-MoLM$\dag$        &  30.33 & 24.47 & \underline{38.48}  & 23.93  & 32.98  & \underline{51.33} \\
&3D-MoLM        &  \textbf{32.00} & \textbf{26.13} & \textbf{40.13}  & \textbf{25.55}  & \textbf{34.64}  & \textbf{52.15} \\
\midrule
\multirow{4}{*}{\lshr{Generalist}} 
&\lshr{Llama2-7B*}        & \lshr{25.22}  & \lshr{21.16} & \lshr{31.48} & \lshr{19.21}  & \lshr{25.22}  & \lshr{43.17} \\
&\lshr{Llama2-7B}        &  \lshr{27.68} & \lshr{22.81} & \lshr{34.73} & \lshr{21.55}  &  \lshr{29.91} & \lshr{46.39} \\
&\lshr{2D-MoLM}        & \lshr{\underline{30.23}} & \lshr{\underline{24.57}} &  \lshr{37.85} & \lshr{22.95}  & \lshr{32.28}  & \lshr{50.08} \\
&\lshr{3D-MoLM$\dag$}        &  \lshr{29.92} & \lshr{24.44} & \lshr{\underline{38.62}}  & \lshr{\underline{22.83}}  & \lshr{\underline{32.30}}  & \lshr{\underline{50.81}} \\
&\lshr{3D-MoLM}        &  \lshr{\textbf{31.81}} & \lshr{\textbf{26.08}} & \lshr{\textbf{40.13}}  & \lshr{\textbf{25.87}}  & \lshr{\textbf{34.99}}  & \lshr{\textbf{51.93}} \\
\bottomrule
\addlinespace[0.1cm]
\end{tabular}
\end{subtable}
\begin{subtable}[t]{\textwidth}
\centering
\caption{A concrete sample of descriptive property QA.}
\vspace{-4pt}
    \begin{tabular}{M{1.25cm} M{2.2cm} M{3cm} M{5.75cm}}
        \toprule
        \scriptsize{Molecule} & Question & Ground truth & 3D-MoLM   \\
        \midrule
        \includegraphics[width=0.075\textwidth]{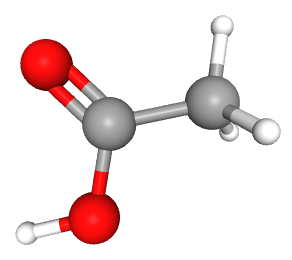} & 
        What are the physical properties of beryllium acetate? & Beryllium acetate is typically a \lsh{white} or colorless \lsh{solid} and is \lsh{soluble in water}. &
        On the physical properties of beryllium acetate, it is a \lsh{white} crystalline \lsh{solid} that is \lsh{soluble in water}. It has a relatively high melting point and boiling point due to the strong intermolecular forces between the beryllium and acetate. \\
        \bottomrule
        \addlinespace[0.1cm]
    \end{tabular}
\end{subtable}
\begin{subtable}[t]{\textwidth}
\centering
\caption{Computed property QA results. We report the MAE with a valid answer rate (\%) -- LMs sometimes fail to follow instructions to generate numerical responses.
We \textcolor{mygreen}{highlight} properties largely dependent on 3D structures.}
{\setlength{\tabcolsep}{1.5pt}
\begin{scriptsize}
\begin{tabular}{llcccccccc}
\toprule
Type& Model &Weight (g/mol)&LogP&TPSA (\text{\AA}$^2$)&Complexity&\textcolor{mygreen}{HOMO (eV)}&\textcolor{mygreen}{LUMO (eV)}&\textcolor{mygreen}{H-L Gap (eV)}&\textcolor{mygreen}{SCF ($10^4$eV)}\\
\midrule
\lshr{Non-LM} & \lshr{Uni-Mol} & \lshr{20.35} &\lshr{0.59}&\lshr{13.48}&\lshr{57.24}&\lshr{0.32}&\lshr{0.35}&\lshr{0.21}&\lshr{0.45}\\
\midrule
\multirow{4}{*}{Specialist}
&Llama2-7B        &  22.10 (96\%) & 1.45 (95\%) & 15.87 (92\%) &  69.74 (93\%) & 1.24 (96\%)  & 1.04 (95\%) & 0.88 (92\%) & 0.70 (99\%)\\
&2D-MoLM        & 21.48 (94\%) & \underline{0.88} (96\%) &  13.52 (92\%) & 55.74 (94\%)  & 0.92 (98\%)  & 0.80 (96\%) & 0.67 (93\%) & 0.71 (99\%)\\
&3D-MoLM$\dag$        & \underline{16.18} (96\%) & 0.95 (96\%) &  \underline{10.26} (94\%) & \underline{49.15} (95\%)  & \underline{0.45} (98\%)  & \underline{0.36} (96\%) & \underline{0.41} (94\%)&\underline{0.39} (99\%)\\
&3D-MoLM        & \textbf{14.79} (95\%) & \textbf{0.66} (97\%) & \textbf{9.71} (93\%) & \textbf{44.85} (94\%) & \textbf{0.26} (97\%) &  \textbf{0.25} (94\%) & \textbf{0.28} (94\%) & \textbf{0.35} (99\%)\\
\midrule
\multirow{5}{*}{\lshr{Generalist}}
&\lshr{Llama2-7B*}        & \lshr{42.18 (82\%)} & \lshr{2.10 (85\%)} & \lshr{27.11 (84\%)} & \lshr{121.87 (76\%)} & \lshr{2.87 (70\%)}  & \lshr{1.89 (71\%)} & \lshr{1.86 (70\%)} & \lshr{3.84 (23\%)}\\
&\lshr{Llama2-7B}        &  \lshr{27.42 (92\%)} & \lshr{1.78 (93\%)} & \lshr{17.07 (90\%)} &  \lshr{78.16 (92\%)} & \lshr{1.89 (90\%)}  & \lshr{1.26 (90\%)} & \lshr{1.25 (91\%)} & \lshr{0.87 (99\%)}\\
&\lshr{2D-MoLM}        & \lshr{20.80 (92\%)} & \lshr{1.36 (94\%)} &  \lshr{12.47 (89\%)} & \lshr{\underline{52.70} (91\%)}  & \lshr{1.52 (93\%)}  & \lshr{1.13 (92\%)} & \lshr{1.09 (88\%)} & \lshr{0.96 (99\%)}\\
&\lshr{3D-MoLM$\dag$}        & \lshr{\underline{19.54} (93\%)} & \lshr{\underline{0.92} (92\%)} &  \lshr{\underline{11.14} (92\%)} & \lshr{54.68 (90\%)}  & \lshr{\underline{0.65} (94\%)}  & \lshr{\underline{0.41} (92\%)} & \lshr{\underline{0.55} (89\%)} &\lshr{\underline{0.49} (99\%)}\\
&\lshr{3D-MoLM}        & \lshr{\textbf{16.58} (92\%)} & \lshr{\textbf{0.78} (95\%)} & \lshr{\textbf{10.90} (90\%)} & \lshr{\textbf{45.49} (89\%)} & \lshr{\textbf{0.35} (95\%)} &  \lshr{\textbf{0.36} (93\%)} & \lshr{\textbf{0.32} (90\%)} & \lshr{\textbf{0.38} (98\%)}\\
\bottomrule
\addlinespace[0.1cm]
\end{tabular}
\end{scriptsize}
}
\label{tab:computed-qa}
\end{subtable}
\begin{subtable}[t]{\textwidth}
\centering
\caption{Concrete samples of computed property QA.}
    \begin{tabular}{M{1.8cm} M{5cm} M{2.65cm} M{2.5cm}}
        \toprule
        \scriptsize{Molecule} & Question & Ground truth & 3D-MoLM   \\
        \midrule
        \includegraphics[width=0.12\textwidth]{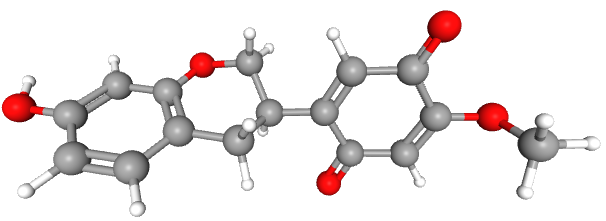} & 
        Please provide the Molecular Weight value for this molecule. If uncertain, provide an estimate. Respond with the numerical value only. & The Molecular Weight for the input molecule is \lsh{286.28 g/mol}. &
        Input molecule has a Molecular Weight of \lsh{288.30 g/mol}. \\
        \midrule
        \includegraphics[width=0.12\textwidth]{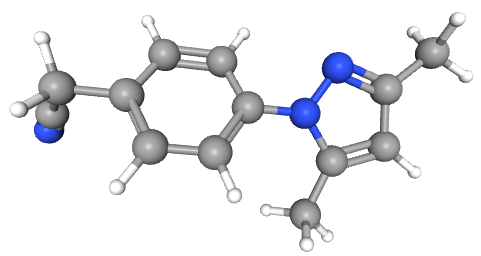} & 
        Could you give me the HOMO-LUMO Gap value of this molecule? If uncertain, provide an estimate. Respond with the numerical value only. & The HOMO-LUMO Gap for the input molecule is \lsh{5.325 eV}. &
        The HOMO-LUMO Gap for the input molecule is \lsh{5.762 eV}. \\
        \bottomrule
    \end{tabular}
\end{subtable}
\label{tab:open-text-qa}
\vspace{-16pt}
\end{table*}
\vspace{-0.1cm}
\section{Conclusion}
\vspace{-0.4cm}
In this work, we introduce 3D-MoLM, a new approach tailored for 3D-molecular language modeling. 
3D-MoLM equips an LM with a 3D molecular encoder for 3D molecule-text interpretation.
This is achieved by a 3D molecule-text projector that aims to map 3D molecular representations into the LM's textual space. 
Furthermore, 3D-MoLM incorporates 3D molecule-centric instruction tuning, enhancing both its adherence to human instructions and 3D molecular understanding. 
Extensive experiments demonstrate that 3D-MoLM excels in various tasks, including molecule-text retrieval, molecule captioning, and open-text molecular QA.

Despite the promising results, our work has a few limitations. Compared with vision-language modeling methods~\citep{instructblip, 3D-LLM}, the scale of our 3D molecule-text dataset is notably constrained, inherently limiting the performance.
This motivates our search for high-quality texts closely related to 3D molecular structures.
Furthermore, this study does not explore other intriguing capabilities of large LMs, such as in-context learning and chain-of-thought reasoning. 
Moreover, integrating principles of invariant learning \citep{InvCF,AdvDrop} can help enhance generalization and mitigate potential hallucinations.
\section{Acknowledgements}
This research is supported by the National Natural Science Foundation of China (92270114) and partially supported by the National Research Foundation Singapore under the AI Singapore Programme (AISG Award No: AISG2-TC-2023-010-SGIL), the Singapore Ministry of Education Academic Research Fund Tier 1 (Award No: T1 251RES2207) and the Google Cloud Research Credits program (Award No: 6NW8-CF7K-3AG4-1WH1).  
This research is also supported by the CCCD Key Lab of Ministry of Culture and Tourism and NExT Research Center.

\bibliography{3D-MoLM.bib}
\bibliographystyle{iclr2024_conference}

\newpage
\appendix
\section{Related Works}\label{sec:related-work}
In this section, we provide a review of literature related to molecule-text modeling, multi-modal instruction tuning, and multi-modal language models.

\textbf{Molecule-Text Modeling.} 
Early research approaches molecular understanding by representing molecules as 1D sequences.
Specifically, KV-PLM~\citep{kvplm} represents molecules using 1D SMILES and employs a masked-language-modeling loss for pretraining on biomedical texts.
MolT5~\citep{molt5} is a T5-based~\citep{T5} model tailored for molecular tasks (\ie SMILES-to-text and text-to-SMILES translations).
Mol-Instruct \citep{mol-ins} utilizes the 1D SELFIES \citep{selfies} molecular descriptor and offers an instruction dataset for biomolecular research. 
Subsequently, 2D molecular graph encoders~\citep{liu2023rethinking} are introduced to explicitly capture topological information.
For example, MoMu~\citep{momu} and MoleculeSTM~\citep{stm} use cross-modal contrastive learning to bridge the representation spaces of molecular graphs and texts. MolCA~\citep{liu2023molca} combines 1D SMILES and 2D graph representations for molecule-to-text generation. ReLM~\citep{ReLM} applies ChatGPT to rerank reaction prediction results proposed by a GNN model.
However, current molecule-text modeling approaches mostly rely on 1D sequences or 2D graphs for molecular representation, overlooking the vital spatial information in 3D molecular conformations. 



\textbf{Multi-Modal Instruction Tuning.} 
Instruction tuning~\citep{FlanModel,flancollection} trains LMs to adhere to natural language instructions, enhancing their ability to generalize to unseen tasks and instructions. 
It has been extended in the multi-modal domain to augment the LM's cross-modal comprehension.
For instance, InstructBLIP~\citep{instructblip} adapts 26 vision-language datasets to the instruction tuning format and introduces instruction-aware visual feature extraction, allowing adaptive and informative feature extraction based on provided instructions.
LLaVA \citep{llava} and MiniGPT-4 \citep{minigpt4} leverage GPT-3.5/GPT-4~\citep{ChatGPT,gpt4} to convert or polish image-text pairs into high-quality instruction-following format, facilitating integrated visual and language understanding.
In the 3D visual domain, 3D-LLM~\citep{3D-LLM} curates a 3D-language instruction tuning dataset to instill a 3D perspective into LMs.
They have showcased the potential of instruction tuning to enhance the cross-modal understanding of LMs. 
However, its application in the 3D molecule-text domain remains largely unexplored.

\textbf{Multi-Modal Language Models.} 
Multi-modal LMs aim to merge the realms of text with other modalities. 
Early vision-language models~\citep{clip,declip} employ contrastive learning to fuse the visual and textual representations. 
Recent studies~\citep{blip2, Flamingo, Frozen, PALM-E} suggest that visual feature spaces can be seamlessly aligned with the input spaces of LMs, enabling them to undertake language generation tasks based on visual inputs. 
Aside from images, researchers have integrated preceptors of diverse modalities, including video \citep{Video-LLaMA}, audio \citep{macawllm} and 3D point clouds \citep{pointllm}, into LMs, giving rise to a series of multi-modal LMs.
We draw inspiration from these revelations to devise 3D-MoLM.

\section{PubChem Dataset and Enrichment Process}\label{app:pubchem-dataset}

\begin{figure}[t]
\centering
\includegraphics[width=0.8\textwidth]{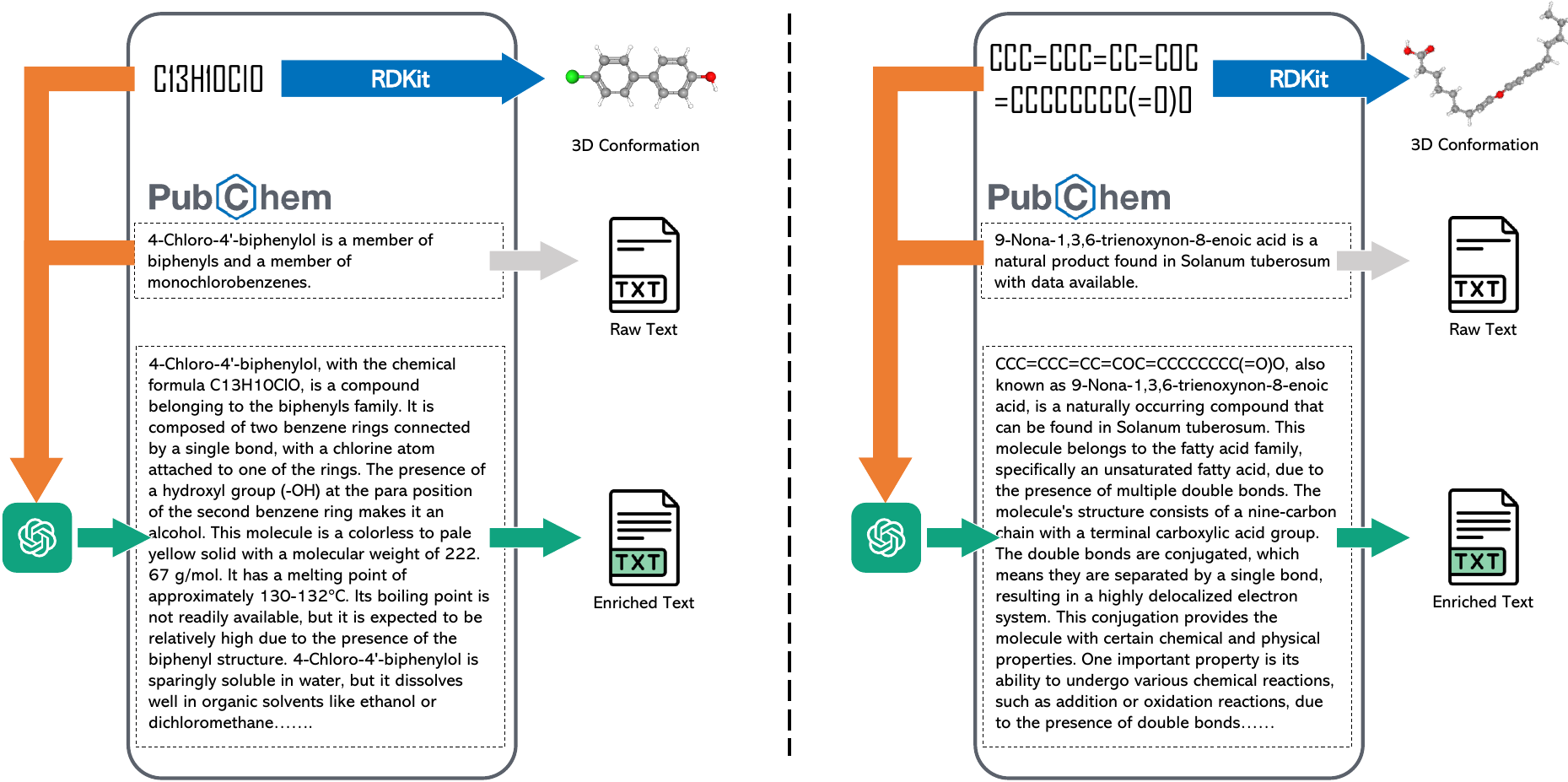}
\caption{Utilizing GPT-3.5 to enrich the descriptions in the pretraining subset of PubChem Dataset.}
\label{fig:pub-enrich}
\vspace{-4pt}
\end{figure}

\textbf{GPT-3.5 Enrichment.}
We begin our data preparation by collecting 324,689 molecular SMILES-text pairs from PubChem~\citep{pubchem}.
However, they include a large number of uninformative texts – the shortest description has only 1 word and $82.29\%$ of the collected molecules have descriptions less than 20 words.
To ensure the effective evaluation for alignment, we first extract a high-quality subset of 15K pairs encompassing descriptions with more than 19 words. 
This curated subset is subsequently partitioned into train / validation / test sets containing 12K / 1K / 2K pairs, respectively.
The less informative remainder is reserved for alignment purpose (\ie stage 1\&2 pretraining).
Yet, we argue that those less informative texts potentially offers an insufficient perspective on molecular properties and structures, which could compromise the 3D molecule-text alignment.  
To address this shortfall, as illustrated in Figure~\ref{fig:pub-enrich}, we employ GPT-3.5 to enhance these molecule descriptions, thus bolstering the less annotated molecules. 
We note that the enrichment process using GPT is exclusively applied to the pretraining set, ensuring that evaluations in downstream tasks (\ie train / validation / test sets) are conducted on original real-world data.

\begin{figure}[t]
\centering
\includegraphics[width=0.85\textwidth]{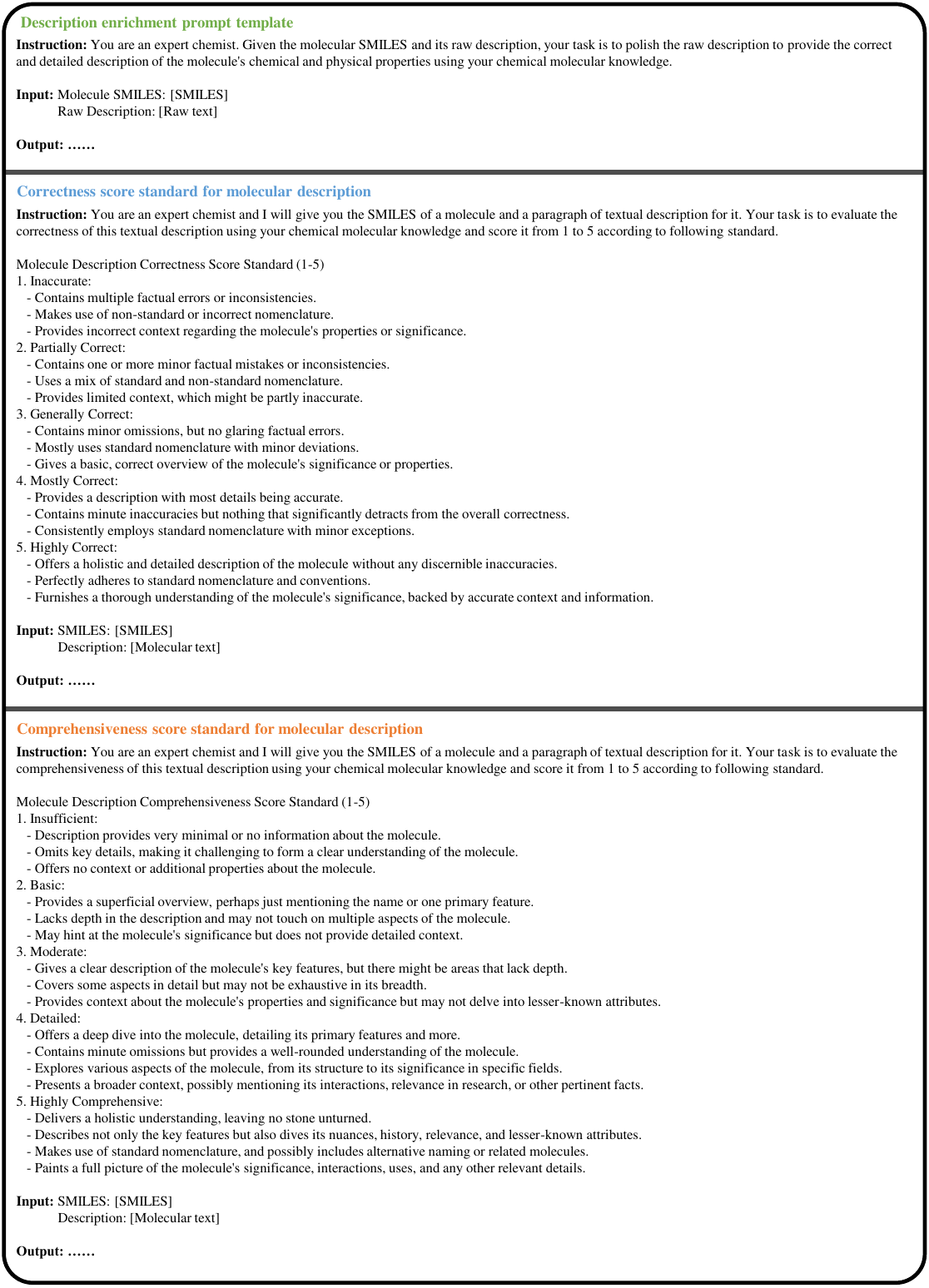}
\vspace{-4pt}
\caption{The prompt template for textual description enrichment and quality evaluation.}
\label{fig:prompt-standard}
\vspace{-12pt}
\end{figure}

\begin{figure}[t]
\centering
\begin{subfigure}[t]{0.4\textwidth}
\includegraphics[width=\textwidth]{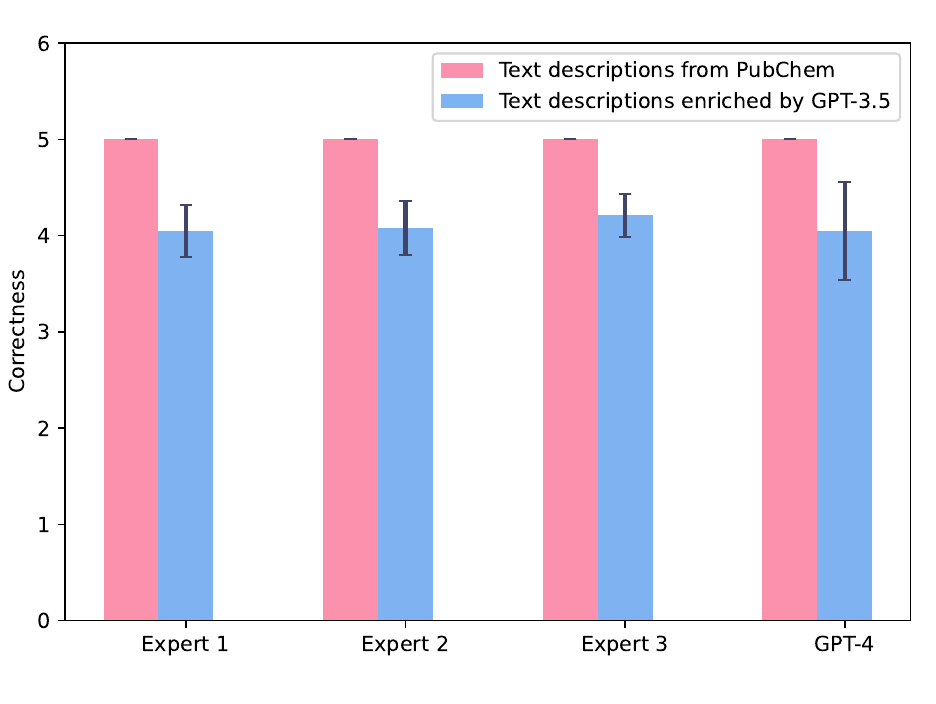}
\caption{Correctness.}
\label{fig:correctness}
\end{subfigure}
\begin{subfigure}[t]{0.405\textwidth}
\includegraphics[width=\textwidth]{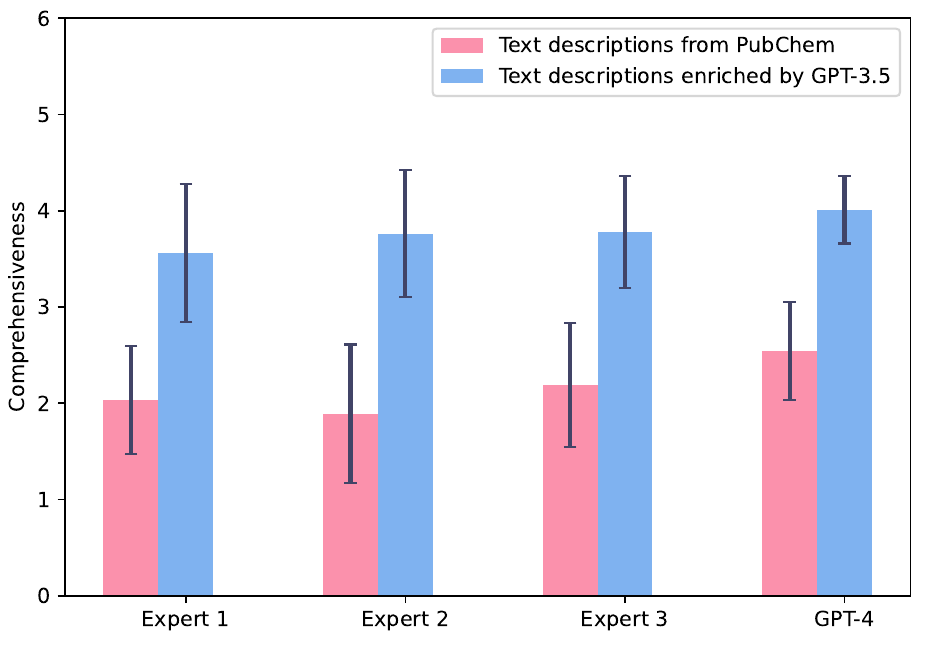}
\caption{Comprehensiveness.}
\label{fig:comprehensiveness}
\end{subfigure}
\vspace{-4pt}
\caption{Correctness and comprehensiveness evaluation for subsets of original PubChem texts and GPT-3.5 enriched ones. Figure~\ref{fig:prompt-standard} shows score standards.}
\label{fig:pubchem-eval}
\vspace{-8pt}
\end{figure}

\textbf{Correctness and Comprehensiveness Evaluation.}
To evaluate the correctness and comprehensiveness of the original texts in PubChem and those enriched by GPT-3.5, we solicited evaluations from three human experts in chemistry and also from GPT-4 with the same score standards in Figure~\ref{fig:prompt-standard}. 
These evaluations, conducted on subsets of 100 (for human experts) and 1,000 (for GPT-4) samples respectively, are presented in Figure~\ref{fig:pubchem-eval}. 
The results indicate that the enhanced descriptions have superior comprehensiveness, while maintaining high correctness.
Adopting the GPT-3.5 generation and subsequent GPT-4 evaluation approach, rather than directly using GPT-4 for text enrichment, was primarily driven by cost considerations; utilizing GPT-4 for the enrichment would have incurred expenses exceeding \$5000.

\begin{table}[t]
\centering
\caption{Number of molecule-text pairs in subsets of PubChem Dataset.}
\vspace{-8pt}
\label{tab:pubchem-dataset}
\scriptsize{
\begin{tabular}{lcccc}\toprule
Subset   & \#Molecule-Text Pair  & \#Min Word & \#Avg Word \\\midrule
Pretrain & 301,658                                & 1               & 17.84              \\
Train    & 12,000                                & 20              & 57.24              \\
Valid    & 1,000                                & 20              & 58.31              \\
Test     & 2,000                               & 20              & 55.21 \\\bottomrule
\end{tabular}
}
\vspace{-12pt}
\end{table}

\textbf{Raw 3D conformation Generation.}
Subsequently, we leverage RDKit to transform the 1D SMILES into raw 3D conformations.
While the raw conformations derived from 1D SMILES might not match the accuracy of costly DFT~\citep{dft} equilibrium conformations, which necessitate several hours for single molecule property calculations, they suffice for preliminary cross-modal alignment.
Meanwhile, RDKit occasionally fails to generate 3D conformations. 
As such, pairs failing this generation process are excluded. 
As shown in Table~\ref{tab:pubchem-dataset}, the resultant dataset split for pretrain/train/validation/test comprises 301,658 / 12,000 / 1,000 / 2,000 3D molecule-text pairs, respectively.
The pretraining set with GPT-enriched descriptions is utilized for 3D molecule-text alignment, detailed in Section~\ref{sec:alignment}. 
Meanwhile, the train, validation, and test sets are used for cross-modal retrieval and molecule captioning tasks, detailed in Section~\ref{sec:retrieval} and ~\ref{sec:captioning}, serving as a measure of alignment quality.

\section{Experiment Setups}\label{app:exp-setup}
In this section, we present the experimental setups for three pretraining stages and downstream tasks.

\textbf{Stage 1 Pretrain.} 
The Q-former attached with a frozen 3D molecular encoder is pertrained for 50 epochs and the number of query tokens in it is set to 8.
AdamW~\citep{adamw} optimizer is adopted with a weight decay of $0.05$ and a learning rate scheduler of a combination of linear warmup with 1000 steps and cosine decay, in which the peak and minimal learning rates are 1e-4 and \lshr{5e-6}, respectively.
And the batch size and maximal text length are 64 and 256, respectively.
The computation overhead is 40 GPU hours on NVIDIA A100 with BFloat16 Mixed precision.

\textbf{Molecule-text retrieval.} 
The Stage-1 checkpoint is finetuned for 10 epochs and evaluated on the downstream subsets (\ie train/validation/test) of PubChem with the same optimizer and learning rate scheduler configurations, \lshr{with the exception of the warmup steps, which are set to 200,} as stage 1 pretraining. 
\lshr{Note that given the relatively small scale of the Q-former and the associated Uni-Mol model, stage 1 does not involve LoRA tuning.}

\textbf{Stage 2 Pretrain.} 
Initiating from the Stage-1 checkpoint, we proceed to the pretraining of Stage 2 for a text generation training of 10 epochs, maintaining the optimizer and scheduler configurations from the prior stage. 
To reduce CUDA memory usage, we integrate LoRA with parameters set to \( \text{r}=8, \alpha=32, \text{dropout}=0.1 \). 
This integration is applied to the modules of \texttt{[k\_proj, v\_proj, q\_proj, o\_proj, gate\_proj, up\_proj, down\_proj]}.
Utilizing this configuration, the resultant LoRA adapter comprises 19M trainable parameters, representing a mere 0.29\% of the total parameter count in Llama2-7B.
With a batch size of 16, maximal token length of 320, and accumulated gradients from 2 batches, the computation overhead is 60 GPU hours on NVIDIA A100 with BFloat16 Mixed precision.

\lshr{\textbf{Stage 3: Generalist Model.}
To obtain one checkpoint for all tasks encompassing molecule captioning, descriptive, non-3D-centric property, and 3D-centric property open-text QA, we mix the training sets of these tasks. 
Given the substantial difference in the size of each dataset, we adopt sampling datasets with probabilities that are proportional to the fourth root of their sizes as in \citep{instructblip}.
Subsequently, the model undergoes training for 400,000 steps with a validation interval of 50,000 steps, utilizing identical optimizer, learning rate scheduler, and LoRA configurations as in stage 2 pretraining. Following this, the checkpoint with the lowest validation loss is evaluated on the test sets of all tasks. The performance of this model is presented in Section \ref{sec:experiment} as a generalist.
}

\textbf{Stage 3: Specialist for Molecule Captioning.} 
The Stage-2 checkpoint is finetuned for 10 epochs and evaluated on the downstream subsets (\ie train/validation/test) of PubChem with the same optimizer, learning rate scheduler, and LoRA configurations as stage 2 pretraining, \lshr{with the exception of the warmup steps, which are set to 200,}. 

\lshr{
\textbf{Stage 3: Specialist for Open-text QA.} 
We possess three specialists for descriptive, non-3D-centric property, and 3D-centric property Open-text-QA, respectively. 
Initially, they universally undergo a comprehensive epoch of all instruction tuning data in 3D-MoIT, employing the same configurations as stage 2, but limited to a single epoch. 
The computational overhead amounts to 200 GPU hours on NVIDIA A100, utilizing BFloat16 Mixed precision. 
Subsequently, the specialists for descriptive, non-3D-centric property and 3D-centric property Open-text-QA are fine-tuned on the sub-dataset of the corresponding task for 5, 5, and 1 epoch, respectively.
}

\begin{figure}[t]
\centering
\includegraphics[width=0.8\textwidth]{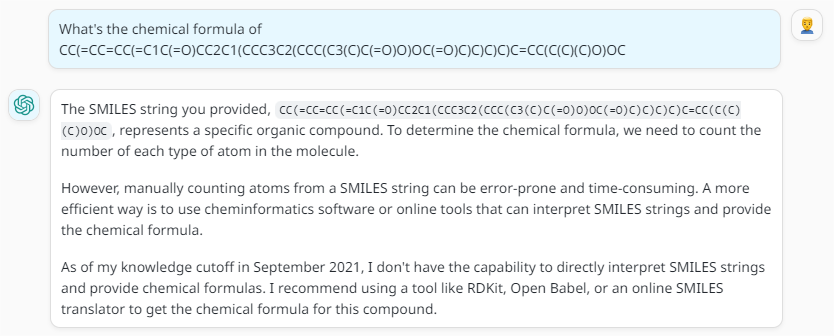}
\caption{\lshr{Existing LLMs (\eg GPT-4) are incapable of accurately counting numbers of atoms and generating chemical formulas of a given SMILES. We use Globostellatic acid B in this example.}}
\label{fig:gpt-count}
\vspace{-12pt}
\end{figure}

\lshr{
\section{Failure Case Study}\label{app:failure-case}
Table \ref{tab:mistake} shows the mistakes 3D-MoLM made in the molecule captioning task, and we offer the subsequent analysis: 

\begin{itemize}[leftmargin=*]
    \item 3D-MoLM incorrectly identifies Globostellatic acid C as Globostellatic acid B, which are structurally distinct solely in the placement of a methoxy branch. As illustrated in Table \ref{tab:failure-case}(b), they have remarkably similar records on PubChem. This suggests that despite accurately identifying primary functions and structures, the discernment of 3D-MoLM at a more refined granularity requires enhancement. We will persist in investigating this in future studies.
    \item 3D-MoLM inaccurately enumerates the number of atoms in Globostellatic acid C as $\text{C}_{34}\text{H}_{48}\text{O}_{6}$, whereas the actual composition is $\text{C}_{33}\text{H}_{48}\text{O}_{7}$. Although this may seem trivial to even non-expert humans, counting atom numbers and providing chemical formulas pose a common challenge for language models. As demonstrated in Figure \ref{fig:gpt-count}, even the powerful GPT-4 model lacks the ability to directly interpret SMILES strings and provide correct chemical formulas. 
\end{itemize}

To rectify these deficiencies, we will persist in refining 3D-MoLM from various perspectives, including 1) integrating a more powerful 3D molecular encoder to distinguish molecules with similar structures, 2) investigating more effective projector structures to bridge the gap between 3D molecular representation and LM space, and 
3) curating more fine-grained datasets for molecule-text alignment and molecule-centric instruction tuning.}

\begin{table*}[t]
\centering
\caption{\lshr{Failure case study.}}
\vspace{-6pt}
\small
\begin{subtable}[t]{\textwidth}
\centering
\caption{\lshr{Case of molecule captioning. Mistakes are highlighted.}}
\label{tab:mistake}
\begin{tabular}{M{3cm} M{4.75cm} M{4.75cm}}
        \toprule
        \scriptsize{Molecule} & Ground Truth & 3D-MoLM  \\
        \midrule
        \includegraphics[width=0.21\textwidth]{figures/cap-molecule-1-v2.png} & 
        \lsh{Globostellatic acid C} is a tricyclic triterpenoid of the isomalabaricane group. It has a role as an antineoplastic agent and a metabolite. It is a tricyclic triterpenoid, an acetate ester, an enone, an ether and an oxo monocarboxylic acid. It is a conjugate acid of a globostellatate C(1-). & 
        \lsh{Globostellatic acid B} is a tricyclic triterpenoid with formula \lsh{C34H48O6}, originally isolated from Globostellata. It has a role as a metabolite. It is a tricyclic triterpenoid, a monocarboxylic acid and an acetate ester. Globostellatic acid B is a natural product found in Globostellata with data available. \\
        \bottomrule
        \vspace{4pt}
    \end{tabular}
\end{subtable}
\begin{subtable}[t]{\textwidth}
\centering
\caption{\lshr{PubChem records of Globostellatic acid C and Globostellatic acid B. Their differences are circled in the 2D molecular graphs below. Here we visualize 2D molecular graphs instead of 3D structures for visual clarity.}}
    \begin{tabular}{M{2.5cm} M{1.5cm} M{2cm} M{6cm}}
        \toprule
        Molecule & Synonym & Molecular Formula & Textual description on PubChem\\
        \midrule
        \includegraphics[width=0.15\textwidth]{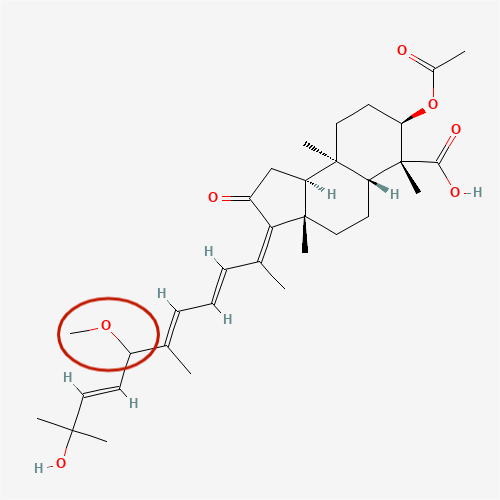} & 
        Globostellatic acid B & $\text{C}_{33}\text{H}_{48}\text{O}_{7}$ &
        Globostellatic acid B is a tricyclic triterpenoid of the isomalabaricane group. It has a role as an antineoplastic agent and a metabolite. It is a tricyclic triterpenoid, an enone, an oxo monocarboxylic acid, an acetate ester and an ether. It is a conjugate acid of a globostellatate B(1-). \\
        \includegraphics[width=0.15\textwidth]{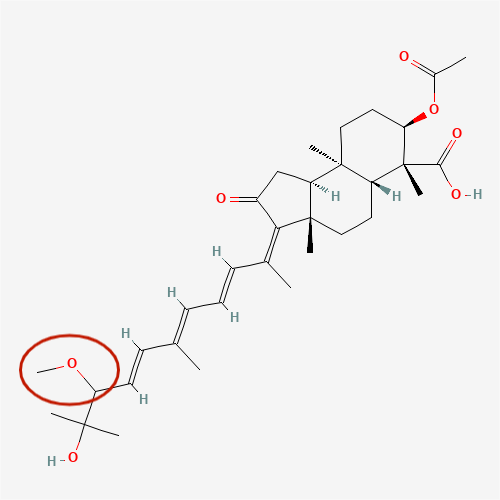} & 
        Globostellatic acid C  & $\text{C}_{33}\text{H}_{48}\text{O}_{7}$ &
        Globostellatic acid C is a tricyclic triterpenoid of the isomalabaricane group. It has a role as an antineoplastic agent and a metabolite. It is a tricyclic triterpenoid, an acetate ester, an enone, an ether and an oxo monocarboxylic acid. It is a conjugate acid of a globostellatate C(1-). \\
        \bottomrule
    \end{tabular}
\end{subtable}
\label{tab:failure-case}
\vspace{-8pt}
\end{table*}

\section{Ablation Study on Input Prompt}\label{app:ablation}
To demonstrate that integrating 3D molecular tokens with 1D SMILES in the prompt enhances molecular understanding, we perform an ablation study and report the results in Table~\ref{tab:ablation}.
We observe that using a combination of both 3D molecular tokens and 1D SMILES as prompts consistently yields superior performance compared to using only one type. 
This can be attributed to that including SMILES ensures the LM taps into its inherent understanding of 1D SMILES. 
Thus, when integrating the 3D molecular encoder and molecule-text projector, the distilled 3D molecular tokens can concentrate on discerning 3D-dependent properties. 
This is akin to enhancing a model with additional features while retaining its existing representation ability, ensuring a more comprehensive understanding.

\begin{table*}[t]
\centering
\caption{Ablation study.
The instruction prompt of 3D-MoLM contains both 3D molecular tokens and 1D SMILES (\cf Figure~\ref{fig:overview}). 
Llama2-7B, going through stage 3 instruction tuning, is prompted with 1D SMILES, while $\ddagger$ denotes prompting with only 3D molecular tokens.
}
\vspace{-4pt}
\begin{subtable}[t]{\textwidth}
\centering
\caption{Molecule captioning on the PubChem Dataset.}
\vspace{-4pt}

\begin{scriptsize}    
\begin{tabular}{lcccccc}
\toprule
Model &  BLEU-2 & BLEU-4 & ROUGE-1 & ROUGE-2 & ROUGE-L & METEOR \\
\midrule
Llama2-7B        &  27.01 & 20.94 & 35.76 & 20.68  &  29.88 & 32.11 \\
3D-MoLM$\ddagger$   & 27.12       & 21.10        & 35.63        & 20.47        & 28.84        & 32.41  \\
3D-MoLM    & 29.25        & 22.07       &36.48       & 21.80        & 30.95        & 33.12  \\
\bottomrule
\addlinespace[0.2cm]
\end{tabular}
\end{scriptsize}

\end{subtable}
\begin{subtable}[t]{\textwidth}
\centering
\caption{Descriptive Property QA results on 3D-MoLM.}
\vspace{-4pt}
\begin{scriptsize} 
\begin{tabular}{lcccccc}
\toprule
Model &  BLEU-2 & BLEU-4 & ROUGE-1 & ROUGE-2 & ROUGE-L & METEOR \\
\midrule
Llama2-7B        &  27.68 & 22.81 & 34.73 & 21.55  &  29.91 & 46.39 \\
3D-MoLM$\ddagger$   &   28.05     &    23.38     &   35.47      &    21.98     &    30.29     &  47.20 \\
3D-MoLM    & 31.61       & 25.88        & 39.93        & 25.68        & 34.78        & 51.74  \\
\bottomrule
\addlinespace[0.2cm]
\end{tabular}
\end{scriptsize}

\end{subtable}
\begin{subtable}[t]{\textwidth}
\centering
\caption{Computed property QA results on 3D-MoLM.}
\vspace{-4pt}
{\setlength{\tabcolsep}{3pt}
\begin{scriptsize}
\begin{tabular}{lcccccccc}
\toprule
Model &Weight (g/mol)&LogP&TPSA (\text{\AA}$^2$)&Complexity&\textcolor{mygreen}{HOMO (eV)}&\textcolor{mygreen}{LUMO (eV)}&\textcolor{mygreen}{H-L Gap (eV)}&\textcolor{mygreen}{SCF ($10^5$eV)}\\
\midrule
Llama2-7B        &  27.42 (92\%) & 1.78 (93\%) & 17.07 (90\%) &  78.16 (92\%) & 1.89 (90\%)  & 1.26 (90\%) & 1.25 (91\%) & 0.87 (99\%)\\
3D-MoLM$\ddagger$        & 24.15 (93\%) & 1.42 (93\%) & 15.02 (92\%) & 58.84 (91\%) & 0.68 (94\%) &  0.52 (93\%) & 0.48 (92\%) & 0.51 (99\%)\\
3D-MoLM        & 16.58 (92\%) & 0.78 (95\%) & 10.90 (90\%) & 45.49 (89\%) & 0.35 (95\%) &  0.36 (93\%) & 0.32 (90\%) & 0.38 (98\%)\\
\bottomrule
\addlinespace[0.2cm]
\end{tabular}
\end{scriptsize}
}
\end{subtable}
\label{tab:ablation}
\vspace{-12pt}
\end{table*}

\lshr{
\section{The effectiveness of 3D perception}
To demonstrate the effectiveness of 3D perception and eliminate the advantage of model scale (\ie Llama2 7B versus T5 780M), we integrate the 3D molecule encoder and the projector into MolT5. 
This facilitates a fair comparison, maintaining the same model size.

\begin{table*}[t]
\centering
\caption{\lshr{Ablation study of 3D perception on MolT5.}}
\begin{scriptsize}    
\begin{tabular}{lccccccc}
\toprule
Base LM & 3D Perception &  BLEU-2 & BLEU-4 & ROUGE-1 & ROUGE-2 & ROUGE-L & METEOR \\
\midrule
\multirow{2}{*}{MolT5-small}
&  \xmark      &  22.53 & 15.23 & 30.44 & 13.45  &  20.30 & 23.98 \\
&    \cmark    & 23.73 & 16.57 &  31.57 & 14.34  & 21.63  & 24.81 \\
\midrule
\multirow{2}{*}{MolT5-base}
&  \xmark      &  24.51 & 16.61 & 32.19 & 14.04  &  21.35 & 26.10 \\
&    \cmark    & 25.55 & 17.31 &  33.36 & 15.49  & 22.62  & 27.54 \\
\midrule
\multirow{2}{*}{MolT5-large}
&  \xmark      &  25.87 & 17.28 & 34.07 & 16.42  &  23.41 & 28.04 \\
&    \cmark    & 26.91 & 18.50 &  35.25 & 17.37  & 25.50  & 29.45  \\
\bottomrule
\addlinespace[0.1cm]
\end{tabular}
\end{scriptsize}
\label{tab:3d-molt5}
\end{table*}

As shown in Table \ref{tab:3d-molt5}, MolT5s equipped with 3D perception consistently surpass those lacking 3D perception across all scales.
This result substantiates our claim that the incorporation of 3D geometry perception enhances the molecular understanding of language models.
}

\section{Limitations and Future Directions}\label{app:limitation}
In this paper, we delve into the domain of 3D molecule-text interpretation, aiming to bridge the understanding between 3D molecular representations and textual descriptions. 
Although the proposed approach demonstrates promising results on various tasks, we recognize the following limitations:

\lshr{\textbf{Expanding to More Molecule-Text Modeling Tasks.} Looking forward, tasks like experimental procedure prediction~\citep{Smiles2actions}, retrosynthesis~\citep{RootSmiles}, and drug-drug interaction~\citep{fang2024moltc} can potentially benefit from 3D molecular-text modeling, but are beyond the scope of this work. This limitation stems partly from the lack of openly available datasets for certain tasks and our aim to maintain a focused scope in this study. Additionally, existing tasks like molecular captioning and molecule-text retrieval can be further developed by introducing new datasets that focus on more complex and novel chemical properties (\eg molecular spectra). Using texts for graph explanation~\citep{li2022let, fangjf2023eva, fangjf2023exgc} and property prediction~\citep{liu2023rethinking} can also be explored. These opportunities for expansion are left for future research.

\textbf{Fine-grained 3D Molecule-Text Alignment.} We include case studies in Appendix~\ref{app:failure-case} showing that 3D-MoLM is limited in discerning fine-grained small molecular structures. Overcoming this issue is crucial for enhancing performance in tasks that demand a precise understanding of 3D molecular structures. One potential direction is to curate a new 3D molecule-text dataset with explicit 3D coordinate references within textual descriptions. Another direction is to explore more powerful 3D molecular encoders and 3D molecule-text projectors.}

\textbf{Dataset Scale.} Other fields of multi-modal modeling have access to datasets comprising more than 10 million paired samples~\citep{cc12m}. 
In contrast, the datasets in this work (\ie PubChem 300K and PubChemQC 3M), are relatively limited in scale. This limitation of dataset scales can impede models from reaching its full potential, and underscores the need for larger and better datasets, especially those with high-quality textual descriptions closely related to 3D molecular structures.

\textbf{Unexplored Capabilities of LMs.} 
Large LMs often exhibit intriguing abilities like in-context learning and chain-of-thought reasoning. 
However, this research does not venture into leveraging these capabilities, primarily due to the lack of specialized datasets detailing step-by-step inferenced.

\textbf{Potential Data Leakage.} 
The evaluation of molecular LMs presents another challenge, particularly with large LMs. 
The vast corpora on which these models are pretrained might inadvertently encompass the test datasets. 
This overlap can introduce biases, thereby inflating performance metrics.

Future work might address these constraints, ensuring a more holistic and accurate understanding of 3D molecule-text interpretation.

\end{document}